\documentclass[10pt]{article} 
\usepackage[preprint]{tmlr}

\usepackage{amsmath,amsfonts,bm}

\def\eqref#1{equation~\ref{#1}}

\def\1{\bm{1}}

\DeclareMathAlphabet{\mathsfit}{\encodingdefault}{\sfdefault}{m}{sl}
\SetMathAlphabet{\mathsfit}{bold}{\encodingdefault}{\sfdefault}{bx}{n}

\usepackage[utf8]{inputenc} 
\usepackage[T1]{fontenc}    
\usepackage{hyperref}       
\usepackage{url}            
\usepackage{booktabs}       
\usepackage{amsfonts}       
\usepackage{nicefrac}       
\usepackage{microtype}      
\usepackage{xcolor}         

\usepackage{natbib} 

\newcounter{myalgorithm}
\renewcommand{\themyalgorithm}{\arabic{myalgorithm}}

\usepackage{makecell}
\usepackage[ruled,vlined]{algorithm2e} 
\usepackage[most]{tcolorbox}
\tcbset{
  colback=gray!5,
  colframe=gray!40!black,
  boxrule=0.4pt,
  arc=1pt,
  left=4pt,
  right=4pt,
  top=2pt,
  bottom=2pt,
  fonttitle=\bfseries,
}
\usepackage{hyperref}
\usepackage{pdfpages}
\usepackage{multirow}
\usepackage{wrapfig}

\usepackage{amsmath}
\usepackage{amssymb}
\usepackage{mathtools}
\usepackage{amsthm}
\usepackage{bbm}
\usepackage[capitalize,noabbrev]{cleveref}

\theoremstyle{plain}
\newtheorem{theorem}{Theorem}[section]

\newtheorem{lemma}[theorem]{Lemma}

\theoremstyle{definition}

\theoremstyle{remark}

\title{Multi-Modal Time Series Prediction via \\Mixture of Modulated Experts}

\author{\name Lige Zhang~\thanks{Part of this work was conducted during his internship visit at Yale University.} \email lige.zhang@emory.edu\\
      \addr Emory University
      \AND
      \name Ali Maatouk \email ali.maatouk@yale.edu \\
      \addr Yale University
      \AND
      \name Jialin Chen \email jialin.chen@yale.edu\\
      \addr Yale University
      \AND
      \name Karthik Charan Konduri \email karthik.charan@gmail.com\\
      \addr Amazon
      \AND
      \name Leandros Tassiulas \email leandros.tassiulas@yale.edu\\
      \addr Yale University
      \AND
      \name Rex Ying \email rex.ying@yale.edu\\
      \addr Yale University
}

\def\month{MM}  
\def\year{YYYY} 
\def\openreview{\url{https://openreview.net/forum?id=XXXX}} 

\begin{document}

\maketitle

\begin{abstract}
    Real-world time series exhibit complex and evolving dynamics, making accurate forecasting extremely challenging. Recent multi-modal forecasting methods leverage textual information such as news reports to improve prediction, but most rely on token-level fusion that mixes temporal patches with language tokens in a shared embedding space. However, such fusion can be ill-suited when high-quality time-text pairs are scarce and when time series exhibit substantial variation in characteristics, thus complicating cross-modal alignment. In parallel, mixture-of-experts (MoE) architectures have proven effective for both time series modeling and multi-modal learning, yet many existing MoE-based modality integration methods still depend on token-level fusion. To address this, we propose \textit{Expert Modulation}, a new mechanism for multi-modal time series prediction that conditions both routing and expert computation on textual signals, enabling direct and efficient cross-modal control over expert behavior. Through theoretical analysis and experiments, our proposed method demonstrates strong improvements in multi-modal time series prediction. The current code implementation is available at \href{https://github.com/BruceZhangReve/MoME}{https://github.com/BruceZhangReve/MoME}.
\end{abstract}

\section{Introduction}
\label{sec:introduction}
Time series prediction plays a critical role in a wide range of domains, including energy~\citep{koprinska2018convolutional, obst2021adaptive}, traffic~\citep{alghamdi2019forecasting, ji2023spatio}, weather~\citep{freeman2018forecasting, lam2023learning}, finance~\citep{yin2021forecasting, ni2023forecasting}, and many others. Traditional time series models, from statistical approaches such as ARIMA~\citep{zhang2003time} to deep learning architectures including MLPs~\citep{zeng2023transformers, chen2023tsmixer}, CNN-based networks~\citep{wu2022timesnet, liu2022scinet}, RNN-style networks~\citep{rangapuram2018deep, hewamalage2021recurrent}, and Transformers~\citep{liu2022pyraformer, zhou2022fedformer, liu2023itransformer}, have achieved success in different domains. More recently, inspired by the effectiveness of mixture-of-experts (MoE) architectures in large language models (LLMs)~\citep{deepseekai2025deepseekv3technicalreport, yang2025qwen3}, several studies~\citep{shi2024time, liu2024moirai} have introduced MoE into time series modeling and achieved promising performance gains.

A common feature of the above methods is that they only use historical sequences for prediction. To that end, they typically struggle when modeling time series in real-world scenarios, such as stock price~\citep{ariyo2014stock, mehtab2020time} and air quality~\citep{zaini2022systematic}, where the time series signal alone can be insufficient to fully capture future variations. In practice, external textual information sources, such as news reports, provide additional predictive cues that are not present in the temporal history itself~\citep{wang2024news}. This has motivated increasing interest in incorporating auxiliary modalities into time series models, a task referred to as multi-modal time series prediction (MMTSP).

Most MMTSP models are built within a common multi-modal learning framework~\citep{alayrac2022flamingo, li2023blip, liu2023visual}: they fuse text tokens and time series tokens (usually temporal patches) in a joint latent space and feed the fused embedding into a pretrained LLM backbone, followed by a downstream regression or classification head for prediction~\citep{jin2023time, jin2024position}, a mechanism we refer to as \textbf{token-fusion}. Nevertheless, a recent work~\citep{zhang2025does} systematically evaluated a wide range of common fusion strategies, including aggregation (add, concat) and projection (MLP, residual) at different fusion locations (early, middle, and late fusion), and found that MMTSP models built on these schemes do not consistently outperform their uni-modal baselines. This suggests that conventional cross-modal interaction methods do not fully exploit the complementary information in textual signals. 

To better understand this limitation, we draw on prior studies~\citep{liu2013modeling, neubauer2024improving} and argue that three plausible reasons explain this observation. First, high-quality time-text pairs are limited, and textual counterparts are often noisy or contain substantial information irrelevant to the underlying time signals. Second, textual descriptions can be insufficiently discriminative to capture the subtle differences between similar time series, leading to ambiguous or inaccurate fusion. Third, time series tokens themselves exhibit high heterogeneity: unbounded value ranges, strong non-stationarity, and substantial variation across samples, making the fusion of time and text tokens extremely challenging. These observations call into question the suitability of token-level fusion for MMTSP, raising the question of \textbf{whether such an approach can adequately model cross-modal interactions, or whether time series and text demand a different integration mechanism.} 

\begin{wrapfigure}{r}{0.52\linewidth}
    \centering
    \includegraphics[width=\linewidth]{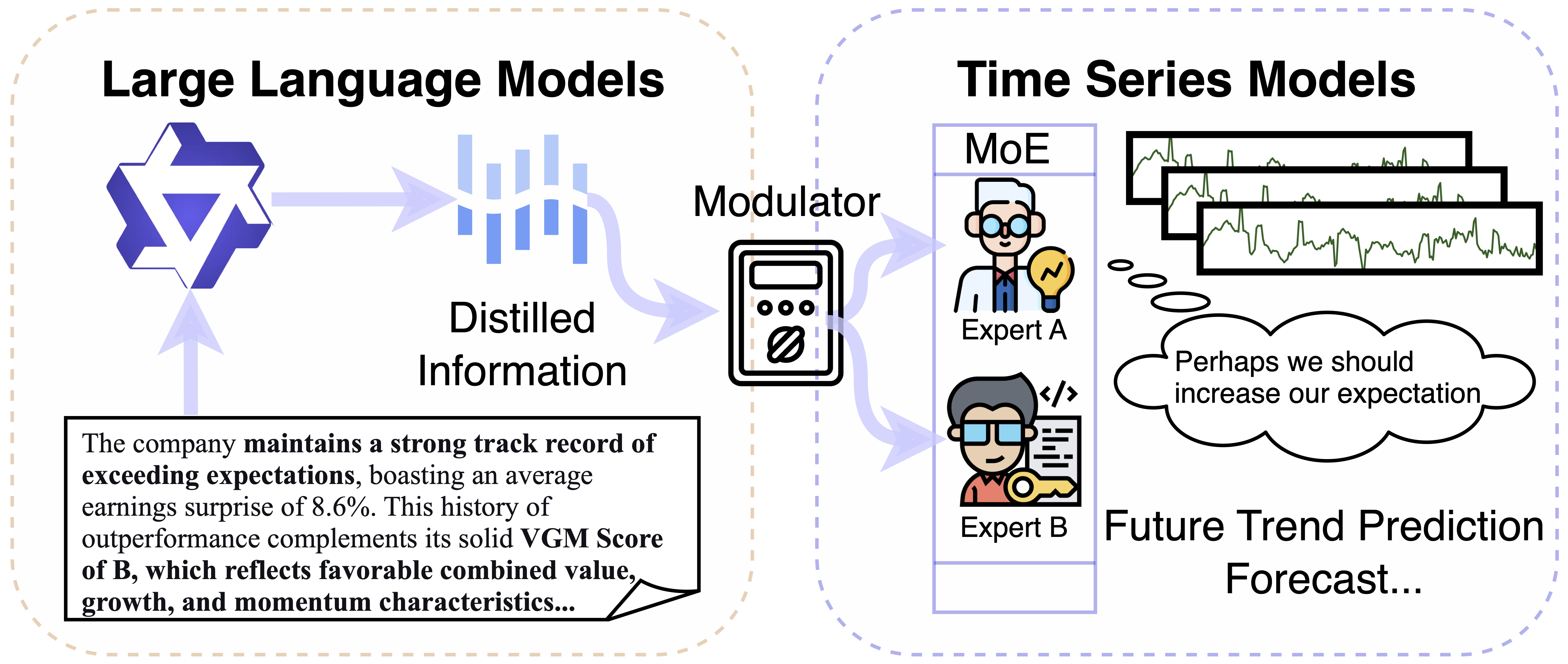}
    \caption{A large language model generates modulation signals that condition both routing and expert computation in a time series MoE model, enabling cross-modal control over temporal experts.}
    \label{fig:intro}
\end{wrapfigure}

To answer this question, we draw inspiration from real-world decision-making systems, where specialized experts adapt their behaviors in response to auxiliary information beyond historical observations, such as epidemiologists who combine infection trajectories with vaccination reports to make predictions. Motivated by this principle, we propose \textbf{Mixture-of-Modulated-Experts} (MoME), a framework that realizes cross-modal interaction via \emph{expert modulation}, in which non-temporal signals directly modulate the computations of temporal experts. Unlike token-level fusion methods that integrate modalities at the representation level, MoME directly injects auxiliary information into expert computations, as shown in Figure~\ref{fig:intro}.

We summarize our main contributions as follows:

(i) We propose \textit{Expert Modulation}, a new mechanism for MMTSP that integrates temporal and textual signals by modulating expert routing and computation within an MoE framework, offering a principled alternative to token-fusion for multi-modal learning. 

(ii) We develop a decomposition view of MoE and theoretically characterize the expressivity of \textit{Expert Modulation} for multi-modal integration.

(iii) We demonstrate the generality and effectiveness of our method for multi-modal time series prediction across diverse backbone architectures, achieving solid improvements over representative baselines.

\section{Related Works}
\label{sec:related_work}

\paragraph{Time Series Prediction via MoE.}
MoE~\citep{jacobs1991adaptive} has become a key architectural design in deep learning models~\citep{lepikhin2021gshard, fedus2022switch}, demonstrating strong performance and efficiency in modern LLMs \citep{deepseekai2025deepseekv3technicalreport, yang2025qwen3}. Recent work has investigated the utilization of MoE in time series models, having successfully developed both small full-shot models~\citep{ni2024mixture, sun2024learning} and large zero-shot foundation models~\citep{shi2024time, liu2024moirai}. However, most studies primarily focus on MoE's potential for uni-modal time series prediction (UMTSP).

\paragraph{Multi-Modal Time Series Prediction.}
According to \citet{zhangdeep2026}, existing MMTSP studies can be broadly categorized into two directions. (i) Self-derived views, where additional modalities are constructed from the time series itself, for example, by converting numerical sequences into images~\citep{liu2024picture, zhang2025timemaster, shen2025multi, zhong2025timevlm} or generating textual descriptions and prompts from temporal observations~\citep{yu2023zero, liu2025timecma}, followed by cross-modal learning with language or vision-language models. (ii) External-information views, where naturally available textual information, such as news reports, are incorporated to provide complementary predictive signals~\citep{jin2023time, jiang2025explainable}. Most methods in the latter category rely on token-fusion, such as concatenation or cross-attention, to establish interactions between temporal and textual representations~\citep{sun2023test, wang2025itformer, quinlan2025chat, zhang2025does, wu2026aurora, lee2026rethinking}. Another line of work employs pre-trained LLMs primarily as expressive sequence encoders for time series~\citep{zhou2023one, 2023arXiv230808469C, tan2024are}, without explicitly incorporating an auxiliary modality for multi-modal integration.


\paragraph{Multi-Modal Learning via MoE.}
MoE has been widely explored for multi-modal learning~\citep{bao2022vlmo, shen2023scaling}. A common design allocates separate, modality-specific parameters to different modalities (e.g., modality-specific experts), which has been shown to benefit deep-layer architectures~\citep{wang2023image}. More recent approaches introduce modality-aware routing or organize experts into modality-specific pools within each MoE layer~\citep{han2024fusemoe, lin2024moma, leim3}. Cross-modal interaction in these methods is still performed at the token (representation) level; however, our approach operates at the expert (function) level.

\section{Preliminary}
\label{sec:preliminary}

\subsection{Multi-Modal Time Series}
Let $\mathbf{X} = [\mathbf{x}_1,\dots,\mathbf{x}_T] \in \mathbb{R}^{T\times D}$ denote a time series with $T$ time steps and $D$ channels, where $\mathbf{x}_t\in\mathbb{R}^{D}$ is the observation at time $t$. We assume the availability of textual information, denoted as $\mathbf{s}_{\mathrm{cxt}}=[s_1,\ldots,s_M]$ (where $M$ is the number of text tokens), which is aligned with the same temporal window and relevant to the prediction target. The goal of MMTSP is to learn a mapping $f:(\mathbf{X},\mathbf{s}_{\mathrm{cxt}})\mapsto Y$, where $Y$ is either (i) a future sequence $[\hat{\mathbf{x}}_{T+1},\dots,\hat{\mathbf{x}}_{T+H}]\in\mathbb{R}^{H\times D}$ over horizon $H$, or (ii) a categorical trend label $Y\in\{1,\dots,N_c\}$ for trend prediction across $N_{c}$ possible classes.

\subsection{Mixture of Experts}
An MoE module consists of a router and a set of expert functions. Let $\mathbf{x}\in\mathbb{R}^{d}$ denote an embedded temporal token (e.g., a temporal segment). For each token, the router computes routing scores
\[
    \mathbf{g}(\mathbf{x}) = \sigma(\mathbf{W}_g\mathbf{x}) = [g_1(\mathbf{x}),\ldots,g_E(\mathbf{x})]^\top \in \mathbb{R}^E,
\]
where $E$ is the number of experts, $\sigma$ is an activation function such as softmax, and $\mathbf{W}_g \in \mathbb{R}^{E \times d}$ is a trainable projection. Each expert $f_e(\cdot)$ is a token-wise transformation, commonly instantiated as a GLU. The MoE output is computed using Top-$K$ sparse routing
\[
    \mathrm{MoE}(\mathbf{x}) = \sum_{i=1}^{E}\lambda_i\,g_i(\mathbf{x})\,f_i(\mathbf{x}), \,\, \text{where} \,\, \lambda_i=\mathbbm{1}[i\in\mathrm{TopK}(\mathbf{g}(\mathbf{x}))], \, \sum_{i=1}^{E}\lambda_i=K,
\]
meaning that exactly $K$ experts are selected for each token.

\section{Mixture of Modulated Experts}
\label{sec:method}
With the above in mind, we first develop a decomposition-view of MoE. Building on this, we introduce our core idea: multi-modal interaction can be achieved by allowing external signals to directly modulate expert behavior within an MoE framework. We then present MoME, a unified framework that instantiates this principle through multiple components.

\subsection{Expert Decomposition and Selective Aggregation}
\label{sec:moe_geoemtric}
To motivate our proposed multi-modal interaction, we first revisit the structure of MoE models. We show that standard feed-forward transformations naturally admit an expert-wise decomposition, and that MoE builds on this structure by selectively controlling the contribution of individual experts.
\begin{lemma}[Decomposition of MLP]
    Let $f(\cdot)$ be a single layer MLP with $E \times C$ hidden units. Then, $f(\cdot)$ can be equivalently written as $f(\mathbf{x}) = \sum_{i=1}^{E} f_i(\mathbf{x})$, where each $f_i(\cdot)$ is a single-layer MLP with $C$ hidden units, corresponding to a disjoint group of hidden neurons.
\label{lemma:decompositoin}
\end{lemma}
Lemma~\ref{lemma:decompositoin}, proved in Appendix~\ref{app:theoretical1}, holds for common variants of MLP (e.g., FFN or GLU). This decomposition establishes a simple connection between dense feed-forward layers and MoE models. In specific, a dense MLP can be regarded as a degenerate mixture in which every component contributes with a fixed coefficient: $\mathrm{MLP}(\mathbf{x}) = \sum_{i=1}^{E} f_i(\mathbf{x})$. In contrast, an MoE generalizes this formulation by introducing input-dependent routing coefficients: $\mathrm{MoE}(\mathbf{x}) = \sum_{i=1}^{E} g_i(\mathbf{x}) f_i(\mathbf{x})$, where $g_i(\mathbf{x})$ controls the contribution of expert $i$. This formulation exposes an important characteristic of MoE. Rather than collapsing all computations into a single representation, MoE maintains an explicit collection of expert-wise components $\left\{ g_i(\mathbf{x})f_i(\mathbf{x})\right\}_{i=1}^{E}$, whose relative contributions can be selectively controlled.

Sparse MoE further exploits this structure by retaining only a subset of expert components. Let $\mathcal{A}$ denote the indices of the $K$ largest routing coefficients, then the sparse routing can be viewed as a selective truncation of expert contributions:
\[
    \mathrm{MoE}_{\mathrm{dense}}(\mathbf{x})
    -
    \mathrm{MoE}_{\mathrm{sparse}}(\mathbf{x})
    =
    \sum_{i\notin\mathcal{A}}
    g_i(\mathbf{x})f_i(\mathbf{x}).
\]
The following result characterizes the magnitude of this truncation.
\begin{theorem}[Selective Truncation in Sparse MoE]
    Let $\{f_i(\mathbf{x})\}_{i=1}^E \subset \mathbb{R}^d$ be the expert directional signals. Assume there exists $B>0$ s.t. $\|f_i(\mathbf{x})\| \le B$ for all $i$, and $|\langle f_i(\mathbf{x}), f_j(\mathbf{x}) \rangle| \le \mu$ for all $i\neq j$. Let $\mathcal{A}$ denote the indices of the $K$ largest routing coefficients from $\{g_i(\mathbf{x})\}_{i=1}^E$. Then, the truncation error
    \[
        \mathcal{L}(\mathbf{x}):= \big\|\mathrm{MoE}_{Dense}(\mathbf{x}) - \mathrm{MoE}_{Sparse}(\mathbf{x})\big\|^2=\Big\|\sum_{i \notin \mathcal{A}} g_i(\mathbf{x}) f_i(\mathbf{x})\Big\|^2,
    \]
    induced by Top-K sparsity is bounded by $\mathcal{L}(\mathbf{x}) \leq [B^2+(E-K-1)\mu]\sum_{i \notin \mathcal{A}}g_i(\mathbf{x})^2$.
\label{theorem:routing}
\end{theorem}
Theorem~\ref{theorem:routing}, proved in Appendix~\ref{app:theoretical2}, shows that sparse routing selectively removes expert contributions with a bounded truncation error. Importantly, the result does not assume that the discarded experts are uninformative. Rather, it formalizes how Top-$K$ routing selectively suppresses a subset of expert contributions instead of treating all experts uniformly.

\begin{wrapfigure}{r}{0.5\linewidth}
    \centering
    \includegraphics[width=\linewidth]{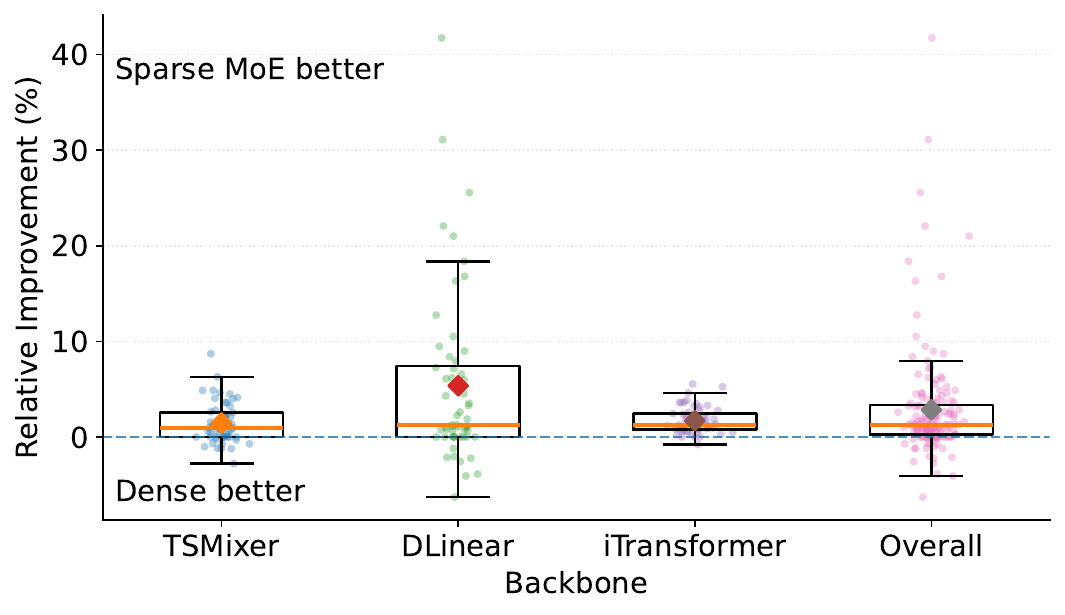}
    \caption{Relative improvement from incorporating sparse MoE into different time series forecasting backbones. Positive values indicate improved performance over the original dense backbone.}
    \label{fig:sparse_moe}
\end{wrapfigure}
This property is particularly relevant to time series modeling. Real-world time series often exhibit heterogeneous local dynamics and observational noise, so that not every computational component is equally useful for every input signal. Sparse expert selection, therefore, provides an architectural mechanism for adapting the set of active transformations to the current temporal pattern. Motivated by this, we examine whether sparse expert activation is useful for time series forecasting. As shown in Figure~\ref{fig:sparse_moe} (detailed in Appendix~\ref{app:ts-moe}), incorporating \textbf{sparse MoE generally improves performance across different dense backbones} and forecasting settings. 

This observation motivates a natural question in the multi-modal setting: if selectively controlling expert contributions is beneficial for time series modeling, can auxiliary information further guide which expert components should contribute to the prediction? We answer this question through \emph{Expert Modulation}.


\subsection{From Selective Expert Aggregation to Expert Modulation}
\label{sec:modulation_beyond}
A standard time series MoE model~\citep{liu2024moirai, shi2024time} determines expert contributions entirely from the temporal input: $F(\mathbf{x}) = \sum_{i=1}^{E} g_i(\mathbf{x})f_i(\mathbf{x})$. When an auxiliary modality $\mathbf{z}$, such as textual context, is available, a common alternative is to first aggregate the temporal expert outputs into $F(\mathbf{x})$ and then condition the resulting representation using $\mathbf{z}$. We refer to this class of operations as \emph{post-aggregation feature modulation}. A representative form~\citep{perez2018film} is 
\[
    F_{\mathrm{fea}}(\mathbf{x}\mid\mathbf{z}) = \gamma(\mathbf{z})F(\mathbf{x}) + \boldsymbol{\beta}(\mathbf{z}),
\]
where $\gamma(\mathbf{z})\in\mathbb{R}$ and $\boldsymbol{\beta}(\mathbf{z})\in\mathbb{R}^{d}$ are context-dependent scales and shifts. Applying MoE decomposition gives
\[
    F_{\mathrm{fea}}(\mathbf{x}\mid\mathbf{z})
    =
    \gamma(\mathbf{z})
    \sum_{i=1}^{E}
    g_i(\mathbf{x})f_i(\mathbf{x})
    +
    \boldsymbol{\beta}(\mathbf{z}).
\]
The limitation of this formulation is that once expert outputs have been aggregated into $F(\mathbf{x})$, their expert-wise structure is no longer accessible to the auxiliary modality. Consequently, the same context-dependent transformation is applied to the entire expert mixture, and the auxiliary signal cannot independently control the relative contributions of different experts.

This is precisely the structure that sparse MoE suggests preserving. If expert-wise selectivity is useful before aggregation, auxiliary information can naturally interact with the model while this decomposition remains accessible, rather than only after individual expert contributions have been mixed. We therefore introduce \emph{Expert Modulation}, which conditions two
complementary aspects of MoE computation: \textbf{which experts contribute} (routing) and \textbf{how each expert contributes} (expert computation).

The first is achieved through context-conditioned routing, $g_i(\mathbf{x}) \rightarrow g_i(\mathbf{x}\mid\mathbf{z})$, which allows auxiliary information to alter expert selection and mixture weights. The second is achieved through expert-specific modulation,
\[
    f_i(\mathbf{x})
    \rightarrow
    f_i(\mathbf{x}\mid\mathbf{z})
    =
    \gamma_i(\mathbf{z})f_i(\mathbf{x})
    +
    \boldsymbol{\beta}_i(\mathbf{z}),
\]
where $\gamma_i(\mathbf{z})\in\mathbb{R}$ and $\boldsymbol{\beta}_i(\mathbf{z})\in\mathbb{R}^{d}$. This allows the same auxiliary context to affect different experts differently. The resulting expert-modulated representation is
\[
    F_{\mathrm{exp}}(\mathbf{x}\mid\mathbf{z})
    =
    \sum_{i\in\mathcal{A}(\mathbf{z})}
    g_i(\mathbf{x}\mid\mathbf{z})
    \left[
        \gamma_i(\mathbf{z})f_i(\mathbf{x})
        +
        \boldsymbol{\beta}_i(\mathbf{z})
    \right],
\]
where $\mathcal{A}(\mathbf{z})$ denotes the context-conditioned Top-$K$ expert set. This formulation extends the principle of selective expert aggregation from the uni-modal to the multi-modal setting. Under this formulation, temporal signals and text tokens jointly decide \emph{whether} an expert contributes and \emph{how} an activated expert contributes. One benefit of this approach is that auxiliary modalities can directly influence expert-wise computation without requiring temporal and textual tokens to be aligned in a shared embedding space.

The difference in where auxiliary information enters the MoE computation also leads to a distinction in model expressivity. \emph{Post-aggregation feature modulation} can only transform the already mixed representation, whereas expert modulation preserves access to individual expert components and can condition them differently before aggregation. As a result, \emph{expert modulation} can represent cross-modal interactions that are inaccessible to post-aggregation feature modulation, as formalized below.
\begin{theorem}[Expressivity of Expert Modulation]
    Assume that the active routing coefficients are normalized, and assume that the routed expert outputs $g_i(\mathbf{x})f_i(\mathbf{x})$ are non-degenerate, in the sense that no non-trivial linear combination of them is constant in $\mathbf{x}$. Then the function class induced by post-aggregation feature modulation is a strict subset of that induced by expert modulation: $\mathcal{F}_{\mathrm{fea}} \subsetneq \mathcal{F}_{\mathrm{exp}}$
    \label{theorem:modulation}
\end{theorem}
Theorem~\ref{theorem:modulation}, proved in Appendix~\ref{app:theoretical3}, formalizes the benefit of preserving expert-wise structure before aggregation. Any global feature modulation can be recovered by applying the same modulation to every expert, whereas expert modulation can additionally apply different context-dependent transformations to different experts.


\begin{figure}
    \centering
    \includegraphics[width=\linewidth]{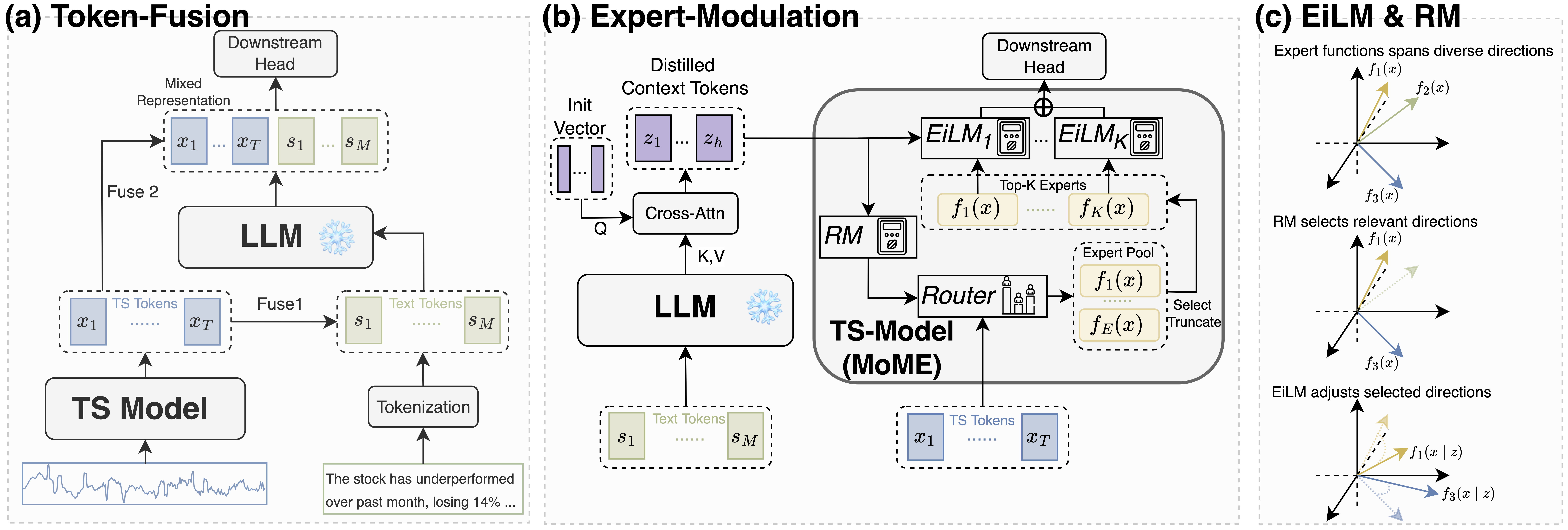}
    \caption{\textbf{(a):} Token-Fusion method, where \textit{Fuse(1)} and \textit{Fuse(2)} denote potential locations that fusion is performed. \textbf{(b):} Expert Modulation method. \textbf{(c):} Geometric intuition on \textit{EiLM} and \textit{RM}.}
    \label{fig:main}
\end{figure}

\subsection{Mixture of Modulated Experts}
\label{sec:mome}
We now integrate the expert modulation principle into MoE-based time series models. The resulting model, MoME, consists of three components. First, cross-attention is used to distill raw textual tokens into compact context tokens. Second, \textit{Router Modulation} (RM) uses the distilled information to adjust the score of a temporal router, thereby changing expert selection and mixture weights. Third, \textit{Expert-independent Linear Modulation} (EiLM) uses the same context to modulate each selected expert output through an expert-specific affine transformation. Figure~\ref{fig:main} (a) describes conventional token-fusion approaches, (b) illustrates the proposed MoME framework, and (c) provides a geometric interpretation of RM and EiLM.

\paragraph{Step 1: Context Token Distillation.}
Given a sequence of language tokens $\mathbf{s}_{cxt} = [s_1, \dots, s_M]$, we obtain their d-dimensional textual representations by applying a linear projection to their final hidden states produced by an LLM
\begin{equation}
    \mathbf{H} = \mathrm{LLM}(\mathbf{s}_{cxt})\,\mathbf{W}_{proj} \in \mathbb{R}^{M \times d}, 
\end{equation}
where $\mathbf{W}_{proj} \in \mathbb{R}^{d'\times d}$ such that $d'$ is the hidden dimension of the LLM. Directly using all tokens for modulation can introduce noise and unnecessary redundancy, while simple pooling may discard informative cues. To obtain a compact yet meaningful textual representation, we adopt a query-based distillation mechanism inspired by BLIP-2~\citep{li2023blip}. Specifically, we introduce $m$ trainable query vectors $\mathbf{Q} \in \mathbb{R}^{m \times d}$, initialized to be mutually orthogonal. Then, a cross-attention is performed, using $\mathbf{Q}$ as queries and $\mathbf{H}$ as keys and values
\begin{equation}
    \mathbf{Z} = \mathrm{SoftMax}(\frac{\mathbf{Q}\mathbf{W}_Q(\mathbf{H}\mathbf{W}_K)^\top}{\sqrt{d}})\mathbf{H}\mathbf{W}_V \, \in \mathbb{R}^{m \times d},
\end{equation}
where $\mathbf{W}_Q, \mathbf{W}_K, \mathbf{W}_V$ denotes learnable projection matrices for queries, keys, and values, respectively. This operation produces $m$ distilled context tokens $\mathbf{Z}$, each representing a high-level semantic summary of textual information. We further obtain a single conditioning vector by pooling
\[
\mathbf{z} = \mathrm{Pool}(\mathbf{Z}), \quad \mathrm{Pool}:\mathbb{R}^{m \times d} \rightarrow \mathbb{R}^{d},
\]
which is used to modulate both the routing and expert computation for the time series MoE model. 

\paragraph{Step 2: Router Modulation.}
For each temporal token $\mathbf{x}\in\mathbb{R}^d$, the router in a time series MoE model with $E$ experts produces an initial routing score $\mathbf{g}(\mathbf{x}) = [g_1(\mathbf{x}), \dots, g_E(\mathbf{x})] \in \mathbb{R}^{E}$. RM then injects a context-dependent shift into the routing score:
\begin{equation}
    \mathbf{g}(\mathbf{x}\mid\mathbf{z}) 
    =
    \mathbf{g}(\mathbf{x})
    +
    \kappa \cdot
    \sigma(\mathbf{W}_{G}\mathbf{z}),
    \quad
    \mathbf{W}_{G}\in\mathbb{R}^{E\times d},
\end{equation}
where $\sigma(\cdot)$ is the same activation used by the router and $\kappa\in[0,1]$ controls the strength of textual modulation. The text-conditioned Top-$K$ mask is then obtained $\lambda_i = \mathbbm{1} \left[i\in\mathrm{TopK}\left(\mathbf{g}(\mathbf{x}\mid\mathbf{z})\right)\right]$. Then, the routing coefficients are re-normalized over the selected Top-$K$ experts.

\paragraph{Step 3: Expert independent Linear Modulation.}
For each expert $f_i(\cdot)$ within the MoE module, we generate an expert-specific scale and bias from the pooled context vector
\begin{equation*}
    \mathbb{I}_i^{\gamma}(\mathbf{z}) = \mathbf{w}_i^\top \mathbf{z},
    \,
    \mathbf{w}_i\in \mathbb{R}^{d}; 
    \,\,\,
    \mathbb{I}_i^{\beta}(\mathbf{z}) = \mathbf{W}_i \mathbf{z}, 
    \,
    \mathbf{W}_i\in \mathbb{R}^{d \times d},
\end{equation*}
The output of the $i$-th expert is then modulated as
\begin{equation}
    f_i(\mathbf{x}\mid\mathbf{z}) = \mathbb{I}_i^{\gamma}(\mathbf{z}) \cdot f_i(\mathbf{x}) + \mathbb{I}_i^{\beta}(\mathbf{z}).
\end{equation}
Finally, the MoME output is obtained by aggregating the modulated expert outputs using the context-conditioned routing weights
\begin{equation}
    \mathrm{MoME}(\mathbf{x} \mid \mathbf{z})
    = \sum_{i=1}^{E} \lambda_i \, g_i(\mathbf{x} \mid \mathbf{z}) \, f_i(\mathbf{x} \mid \mathbf{z}),
    \label{eq:MoME}
\end{equation}
In this design, textual signals do not interact directly with temporal tokens. Instead, they enter the MoE by modulating expert behavior through per-expert affine transformations and context-conditioned routing. In contrast, most existing approaches rely on token-fusion, where temporal tokens are projected and fused with textual tokens in a shared embedding space.

\subsection{Complexity Analysis}

Let $N$ be the number of temporal tokens, $M$ the number of text tokens, $m$ the number of distilled context tokens, $d$ the model dimension, $\tilde{d}$ the expert hidden dimension, $E$ the number of experts, and $K$ the number of activated experts. We exclude the cost of the pre-trained LLM encoder and consider the additional computation after obtaining the projected textual representations $\mathbf{H}\in\mathbb{R}^{M\times d}$. The uni-modal sparse MoE backbone has complexity $\mathcal{O}(NEd + NKd\tilde{d})$, where the two terms correspond to routing and activated expert computation, respectively. MoME additionally performs context-token distillation through cross-attention between $m$ learnable queries and $M$ textual tokens, requiring $\mathcal{O}((M+m)d^2 + mMd)$ computation. RM adds only lower-order computation, while EiLM introduces an additional $\mathcal{O}(Ed^2)$ cost. Thus, the overall post-LLM complexity is $\mathcal{O}\left((M+m)d^2+mMd+NEd+NKd\tilde d+Ed^2\right)$, which reduces to $\mathcal{O}\left(Md^2+NEd+NKd\tilde d+Ed^2\right)$ for a small fixed $m$. In contrast, direct temporal-text cross-attention
requires $\mathcal{O}\left((N+M)d^2+NMd\right)$ computation. Since $m\ll N$ and $E$ are small in practice, MoME incurs only moderate additional computational cost for multi-modal interaction.

\subsection{Expert Modulation on Different Backbones}
\label{sec:timemoe_modulated}
Expert Modulation defines a general, backbone-agnostic principle for multi-modal time series modeling. To demonstrate its flexibility, we instantiate this framework in three representative architectures with increasing modeling complexity: a linear model (MMLinear), an MLP-based model (MoME), and a transformer-based model (MiTransformer). 

(i) \textbf{MoME} is the primary instantiation of our framework. The input time series is segmented into patches. Experts are implemented as MLPs with expert modulation applied.

(ii) \textbf{MMLinear} is a lightweight variant that operates directly on the full sequence, where each expert is reduced to a linear transformation with expert modulation applied. 

(iii) \textbf{MiTransformer} is a multi-modal extension of iTransformer that replaces FFN blocks with MoME modules. It also operates on the full sequence without patching.

\section{Experiments}
\label{sec:experiments}

\subsection{Experiments Setup}
\label{sec:setup}

\paragraph{Datasets.}

Our primary evaluation focuses on MMTSP, with MT-Bench~\citep{chen2025mtbench} serving as the main benchmark because its textual information is relatively well aligned with and the corresponding time series, providing a desirable setting for evaluating cross-modal integration. We evaluate MoME on its weather and finance domains. To assess generality beyond this primary setting, we include complementary experiments on additional real-world datasets from Time-IMM~\citep{chang2026timeimm}, as well as the Time-MMD benchmark~\citep{liu2024time}. In addition, to examine the theoretical insights (Theorem~\ref{theorem:routing}) of MoE and its role in time series modeling, we conduct supporting experiments on seven UMTSP benchmarks~\citep{lai2018modeling, zhou2021informer, wu2021autoformer} (see Appendix~\ref{app:uni_results}). Finally, we extend the evaluation to multi-channel multi-modal scenarios using the MT-Bench and Time-IMM benchmarks. Detailed dataset descriptions are provided in Appendix~\ref{app:data1}.


\paragraph{Baselines.}
We consider two types of baselines: (a) those that do not utilize external text and (b) those that leverage external text for prediction. For type (a), we include four supervised time series forecasting models: PatchTST~\citep{nie2022time}, iTransformer~\citep{liu2023itransformer}, TS-Mixer~\citep{chen2023tsmixer}, DLinear~\citep{zeng2023transformers}, and a pre-trained time series foundation model (TSFM), TimeMoE~\citep{shi2024time}. We further include GPT4TS~\citep{zhou2023one} and Time-LLM~\citep{jin2023time}, which employ pre-trained language models as time series encoders. For the pre-trained models, the backbone parameters are frozen, and only the downstream prediction head is trained. For type (b), many existing multi-modal time series models may not be directly comparable, as they follow different modeling mechanisms such as agentic forecasting~\citep{jiang2025explainable, wang2024news} or temporal reasoning~\citep{quinlan2025chat, wang2025itformer, zhang2025timemaster}. We therefore follow~\citet{zhang2025does} and implement three token-fusion baselines: MLP (MoE)+LM, TimeMoE+LM, and DLinearP+LM, where ``+LM'' denotes token-level incorporation of external text. MLP (MoE) uses MLP experts, while DLinearP is a patched variant of DLinear. These baselines provide a controlled comparison between expert modulation and conventional token-fusion strategies.

\paragraph{Settings.}
To ensure a fair comparison, we standardize key hyperparameters (e.g., hidden dimensions and number of layers) and align training configurations across all experiments. All models are implemented in PyTorch and trained on a single NVIDIA RTX A6000 (48GB) GPU. For text processing in the main experiments, we use a pre-trained LLM, QwenMoE\_A2.7B, and truncate the input text to a maximum length of 2048 tokens. Detailed experimental setups are provided in the Appendix~\ref{app:MoME_additional_tableseter_main}.


\begin{table}[]
    \centering
    \caption{Model performance on MT-Bench and Time-IMM. Red bold denotes the best result, and bold denotes the second-best result. \textit{Imp.} (\%) is computed relative to the strongest baseline. Uni-channel results use denormalized metrics, while multi-channel results use normalized metrics.}
    \resizebox{\linewidth}{!}{
        \begin{tabular}{ll|cccccc|cccc|c}
\toprule
\textbf{Task} & \textbf{Metric} &
\textbf{PatchTST} &
\textbf{DLinear} &
\textbf{TS-Mix} &
\textbf{Time-MoE} &
\textbf{Time-LLM} &
\textbf{GPT4TS} &
\makecell{\textbf{Time-MoE}\\\textbf{+LM}} &
\makecell{\textbf{DLinearP}\\\textbf{+LM}} &
\makecell{\textbf{MLP(MoE)}\\\textbf{+LM}} &
\makecell{\textbf{MoME}\\\textbf{(ours)}} &
\textit{Imp.(\%)}\\
\midrule

\multicolumn{12}{l}{\textbf{Uni-Channel: Forecast}} \\
\multirow{2}{*}{Finance (Sort)} 
& MAPE$\downarrow$  & 2.774 & 2.863 & 2.759 & 3.419 & 2.898 & 2.944 & 2.772 & 2.805 & \textcolor{red}{\textbf{2.545}} & \textbf{2.605} & -2.357 \\
& MAE$\downarrow$   & 3.264 & 3.383 & 3.220 & 3.984 & 3.411 & 3.540 & 3.248 & 3.419 & \textbf{3.136} & \textcolor{red}{\textbf{3.058}} & 2.487 \\
\multirow{2}{*}{Finance (Long)} 
& MAPE$\downarrow$  & 3.832 & 3.853 & 3.802 & 4.564 & 4.328 & 4.272 & \textbf{3.749} & 4.010 & 3.757 & \textcolor{red}{\textbf{3.523}} & 6.028 \\
& MAE$\downarrow$   & 3.604 & 3.620 & 3.553 & 4.326 & 4.023 & 3.886 & \textbf{3.427} & 3.606 & 3.523 & \textcolor{red}{\textbf{3.228}} & 5.807 \\

\multirow{2}{*}{Weather (Short)} 
& MSE$\downarrow$   & 10.448 & 10.438 & 18.222 & 21.635 & 10.658 & 11.947 & \textbf{10.211} & 10.374 & 11.233 & \textcolor{red}{\textbf{10.020}} & 1.871  \\
& MAE$\downarrow$   & 2.404  & 2.449  & 3.292  & 3.572  & 2.453  & 2.537  & \textbf{2.397}  & 2.407  & 2.508  & \textcolor{red}{\textbf{2.389}} & 0.333  \\
\multirow{2}{*}{Weather (Long)} 
& MSE$\downarrow$   & 14.312 & 13.392 & 24.445 & 40.887 & \textbf{12.805} & 13.901 & 14.894 & 13.666 & 13.331 & \textcolor{red}{\textbf{11.823}} & 7.669  \\
& MAE$\downarrow$   & 2.857  & 2.765  & 3.846  & 5.010  & \textbf{2.738}  & 2.812  & 2.923  & 2.809  & 2.771  & \textcolor{red}{\textbf{2.620}} & 4.310  \\

\multirow{2}{*}{EPA-Air} 
& MSE$\downarrow$  & 19.372 & 20.964 & 23.706 & 20.135 & 16.960 & 18.938 & 20.135 & 30.745 & 31.612 & \textcolor{red}{\textbf{16.148}} & 4.788 \\
 & MAE$\downarrow$  & 3.231 & 3.354 & 3.677 & 3.593 & 3.060 & 3.256 & 3.586 & 4.141 & 4.227 & \textcolor{red}{\textbf{2.992}} & 2.222 \\
 \multirow{2}{*}{ILINet} 
& MSE$\downarrow$  & 1.820 & 1.549 & 1.573 & 0.863 & 0.905 & 1.602 & 0.868 & 3.819 & \textbf{0.824} & \textcolor{red}{\textbf{0.811}} & 1.578 \\
 & MAE$\downarrow$  & 0.993 & 0.845 & 0.816 & 0.589 & \textbf{0.561} & 0.679 & 0.570 & 1.915 & 0.555 & \textcolor{red}{\textbf{0.547}} & 1.441 \\
 
\midrule
\multicolumn{12}{l}{\textbf{Uni-Channel: Trend}} \\
\multirow{2}{*}{Finance (Short)} 
& 3-way$\uparrow$   & 39.674 & 38.044 & 47.826 & 42.391 & 40.761 & 48.913 & 46.739 & \textbf{49.315} & 43.478 & \textcolor{red}{\textbf{66.849}} & 17.534  \\
& 5-way$\uparrow$   & 40.217 & 29.348 & \textbf{42.391} & 38.044 & 38.587 & 40.217 & 35.326 & 39.041 & 40.127 & \textcolor{red}{\textbf{55.978}} & 13.587 \\
\multirow{2}{*}{Finance (Long)} 
& 3-way$\uparrow$   & \textbf{50.802} & 41.096 & 50.685 & 44.178 & 49.315 & 48.287 & 48.630 & 45.109 & 48.288 & \textcolor{red}{\textbf{63.699}} & 13.014  \\
& 5-way$\uparrow$   & 41.438 & 30.822 & \textbf{44.521} & 36.645 & 42.123 & 41.438 & 30.822 & 41.848 & 33.219 & \textcolor{red}{\textbf{51.027}} & 6.506  \\

\multirow{2}{*}{Weather (Short)} 
& Past$\uparrow$    & \textcolor{red}{\textbf{93.877}} & 69.231 & 91.484 & 85.440 & 87.918 & 88.189 & \textbf{92.582} & 89.011 & 91.758 & 91.484 & -2.393  \\
& Future$\uparrow$  & 51.374 & 50.824 & 53.022 & \textbf{53.571} & 42.674 & 53.297 & 42.307 & 42.582 & 43.132 & \textcolor{red}{\textbf{55.220}} & 1.649  \\
\multirow{2}{*}{Weather (Long)} 
& Past$\uparrow$    & 89.973 & 82.656 & 90.786 & 89.431 & 84.553 & 85.366 & \textbf{91.599} & 88.347 & 86.450 & \textcolor{red}{\textbf{93.496}} & 1.897  \\
& Future$\uparrow$  & 26.558 & 49.865 & 48.510 & 48.510 & \textbf{50.949} & 50.407 & 42.005 & 42.276 & 41.192 & \textcolor{red}{\textbf{53.388}} & 2.439 \\
\midrule

\multicolumn{12}{l}{\textbf{Multi-Channel: Forecast}} \\
\multirow{2}{*}{Weather} 
& MSE$\downarrow$  & 1.352 & 1.385 & \textcolor{red}{\textbf{1.307}} & 1.627 & 1.379 & 1.388 & 1.463 & 1.331 & 1.340 & \textbf{1.326} & -1.454 \\
 & MAE$\downarrow$  & 0.846 & 0.854 & \textbf{0.837} & 0.969 & 0.849 & 0.858 & 0.899 & 0.855 & 0.858 & \textcolor{red}{\textbf{0.835}} & 0.239 \\
\multirow{2}{*}{Gdelt} 
& MSE$\downarrow$  & \textbf{1.762} & 1.877 & 1.809 & 1.808 & 1.770 & 1.766 & 1.825 & 1.899 & 1.960 & \textcolor{red}{\textbf{1.760}} & 0.114 \\
 & MAE$\downarrow$  & 0.874 & 0.934 & 0.887 & 0.873 & 0.875 & \textcolor{red}{\textbf{0.872}} & 0.892 & 0.937 & 0.950 & \textbf{0.878} & -0.688 \\
\bottomrule
\end{tabular}
    }
    \label{tab:main}
\end{table}

\subsection{Effectiveness for Prediction}
\label{sec:effectiveness}

We evaluate MoME under both uni-channel and multi-channel settings. The main results on real-world datasets (MT-Bench and Time-IMM) are reported in Table~\ref{tab:main}. Additional results on Time-MMD are provided in Appendix~\ref{app:mome,time-mmd}.

\paragraph{Uni-Channel.}
MoME achieves strong overall performance across datasets, with particularly clear gains on finance trend prediction and competitive or improved results on most forecasting tasks. This indicates that textual information is particularly effective in helping time series models capture directional or trend-level signals, compared to precise numerical values. One possible explanation is that textual reports are inherently coarse-grained and often summarize high-level dynamics (e.g., market sentiment or macro trends), making them better aligned with coarse-resolution targets such as trend prediction. Overall, these results demonstrate that MoME can effectively extract useful information from complex texts (e.g., financial reports as shown in Figure~\ref{fig:main_case}) and improve prediction.



\paragraph{Multi-Channel Setting.}
To enable multi-modal, multi-channel prediction, we adopt a channel-independent strategy, where each variable is modeled separately and textual information is shared across channels. MoME consistently outperforms most baselines, although the gains are more moderate. This is likely because part of the predictive signal can already be captured through cross-channel interactions within the time series modality itself, which reduces the marginal gain of textual information. Nevertheless, the results demonstrate that expert modulation integrates seamlessly with multi-channel scenarios and achieves strong performance. 

\begin{table}[]
    \centering
    \caption{Ablation on expert modulation across backbones and tasks. Bold numbers indicate the best results within each backbone family. Avg. Imp. (\%) denotes the average relative improvement of the best modulated variant over the corresponding uni-modal case ($w/o$\, EM).}
    \resizebox{\linewidth}{!}{
        \begin{tabular}{ll
                ccc|ccc|ccc|c}  
\toprule
\multirow{2}{*}{Task} & \multirow{2}{*}{Metric}
& \multicolumn{3}{c}{MoME}
& \multicolumn{3}{c}{MMLinear}
& \multicolumn{3}{c}{MiTransformer}
& \multirow{2}{*}{Avg. Imp. (\%) } \\   
\cmidrule(lr){3-5}\cmidrule(lr){6-8}\cmidrule(lr){9-11}
& 
& $w/o$\, EM & $w/\,$EiLM & $w/\,$EM
& $w/o$\, EM & $w/\,$EiLM & $w/\,$EM
& $w/o$\, EM & $w/\,$EiLM & $w/\,$EM
&  \\ 
\midrule

\multicolumn{12}{l}{\textbf{Uni-Channel: Forecast}} \\
\multirow{2}{*}{Finance (Short)} & MAPE$\downarrow$
& 2.670 & \textbf{2.659} & 2.671
& 2.856 & 2.819 & \textbf{2.789}
& 2.830 & 2.885 & \textbf{2.797}
& 1.308 \\
& MAE$\downarrow$
& 3.207 & \textbf{3.058} & 3.131
& 3.392 & 3.311 & \textbf{3.261}
& 3.561 & 3.602 & \textbf{3.497}
& 3.435 \\

\multirow{2}{*}{Finance (Long)} & MAPE$\downarrow$
& 3.758 & 3.531 & \textbf{3.523}
& 4.208 & 3.993 & \textbf{3.913}
& 3.789 & 3.730 & \textbf{3.676}
& 5.415 \\
& MAE$\downarrow$
& 3.505 & 3.253 & \textbf{3.228}
& 3.942 & 3.594 & \textbf{3.572}
& \textbf{3.375} & 3.380 & 3.389
& 5.714 \\

\multirow{2}{*}{Weather (Short)} & MSE$\downarrow$
& 10.673 & \textbf{10.020} & 10.225
& 10.837 & 10.709 & \textbf{10.585}
& 12.213 & 15.192 & \textbf{11.011}
& 6.095 \\
& MAE$\downarrow$
& 2.480 & \textbf{2.389} & 2.431
& 2.499 & 2.476 & \textbf{2.460}
& 2.719 & 2.892 &\textbf{2.469}
& 4.808 \\

\multirow{2}{*}{Weather (Long)} & MSE$\downarrow$
& 13.580 & 13.337 & \textbf{12.262}
& 13.577 & 13.424 & \textbf{12.381}
& 16.360 & \textbf{15.263} & 15.337
& 8.407 \\
& MAE$\downarrow$
& 2.785 & 2.756 & \textbf{2.666}
& 2.778 & 2.762 & \textbf{2.681}
& 3.093 & \textbf{2.957} & 2.999
& 4.054 \\

\multicolumn{12}{l}{\textbf{Uni-Channel: Trend}} \\
\multirow{2}{*}{Finance (Short)} & 3-way$\uparrow$
& 45.108 & \textbf{66.849} & 63.587
& 45.652 & 45.108 & \textbf{47.283}
& 36.413 & 40.217 & \textbf{44.565}
& 10.508 \\
& 5-way$\uparrow$
& 42.935 & 53.261 & \textbf{55.978}
& 40.761 & \textbf{46.739} & 45.109
& 29.891 & 23.913 & \textbf{33.152}
& 7.427 \\

\multirow{2}{*}{Finance (Long)} & 3-way$\uparrow$
& 46.233 & 60.616 & \textbf{63.699}
& 44.521 & \textbf{58.904} & 52.397
& 49.658 & 48.289 & \textbf{52.055}
& 11.415 \\
& 5-way$\uparrow$
& 40.411 & \textbf{51.027} & 48.973
& 41.438 & \textbf{47.603} & 45.109
& 41.781 & 42.123 & \textbf{43.836}
& 6.279 \\

\multirow{2}{*}{Weather (Short)} & Past$\uparrow$
& 87.363 & \textbf{91.484} & 90.786
& 82.967 & 83.242 & \textbf{83.517}
& 88.189 & 87.088 & \textbf{89.011}
& 1.831 \\
& Future$\uparrow$
& 52.198 & 53.022 & \textbf{54.121}
& 50.275 & 51.923 & \textbf{52.747}
& \textbf{51.374} & 48.626 & 51.099
& 1.373 \\

\multirow{2}{*}{Weather (Long)} & Past$\uparrow$
& 87.534 & \textbf{93.496} & 92.141
& 85.637 & 87.805 & \textbf{88.347}
& 87.263 & 85.637 & \textbf{87.805}
& 3.071 \\
& Future$\uparrow$
& 49.594 & 52.575 & \textbf{53.388}
& 48.781 & \textbf{51.762} & 49.323
& 47.426 & 47.968 & \textbf{48.510}
& 2.620 \\

\bottomrule
\end{tabular}
    }
    \label{tab:ablation}
\end{table}

\subsection{Ablation Studies}
\label{sec:Ablation}

\paragraph{Component Effect.} 
To assess the effectiveness of Expert Modulation, we conduct component-level ablation studies by progressively introducing its two components while keeping all other hyperparameters identical. Specifically, we evaluate three backbone variants of the framework: MoME, MMLinear, and MiTransformer. For each backbone, we consider three settings: a purely uni-modal baseline without textual input, denoted as ($w/o$ EM); a multi-modal variant that introduces textual information only through Expert-independent Linear Modulation ($w/$ EiLM); and the full Expert Modulation variant ($w/$ EM), where EM consists of EiLM together with Router Modulation (RM). Therefore, the comparison from $w/o$ EM to $w/$ EiLM directly measures the benefit of incorporating textual information through expert computation, while the comparison from $w/$ EiLM to $w/$ EM isolates the additional contribution of text-conditioned routing. The results in Table~\ref{tab:ablation} show that incorporating EiLM consistently improves performance over most uni-modal baselines across all backbone families, while adding RM further improves performance in 67\% of the scenarios. Notably, MoME, despite using a simple time series backbone, often outperforms MiTransformer. This suggests that in the scenarios we study, effective cross-modal integration can substantially enhance predictive performance even without highly complex time series architectures.

\begin{figure}
    \centering
    \includegraphics[width=\linewidth]{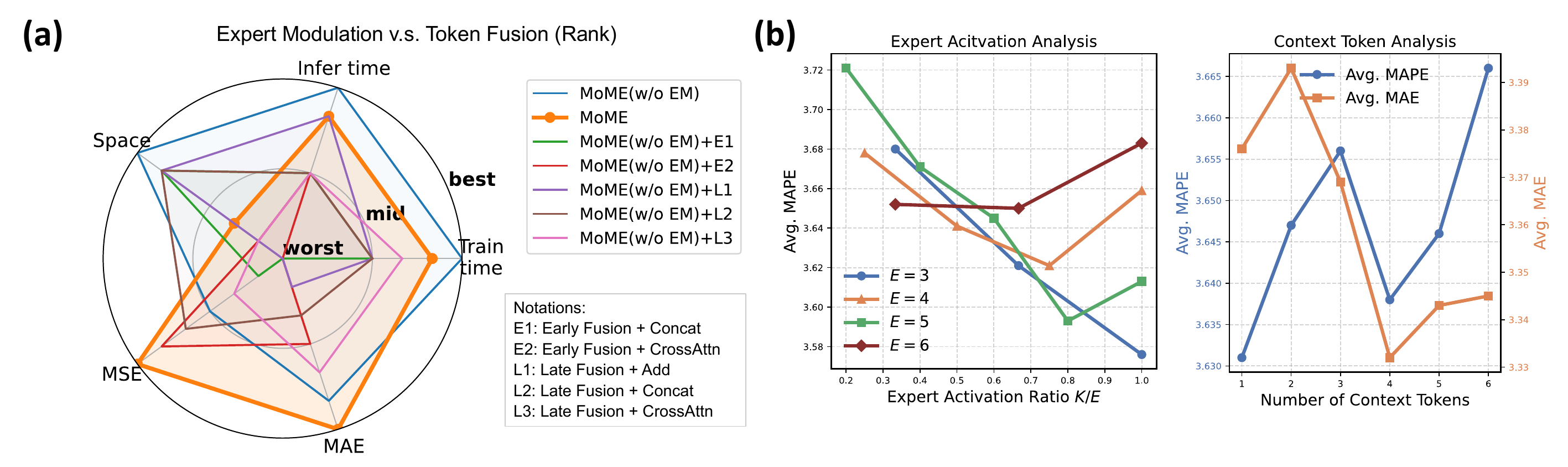}
    \caption{\textbf{(a):} token fusion v.s. expert modulation. \textbf{(b):} hyperparameter analysis. Detailed metric numbers for generating those figures are provided in Appendix~\ref{app:fusion-compare} and Appendix~\ref{app:MoME_additional_tables}.}
    \label{fig:hyperparameter_fig}
\end{figure}
\begin{figure}
    \centering
    \includegraphics[width=0.95\linewidth]{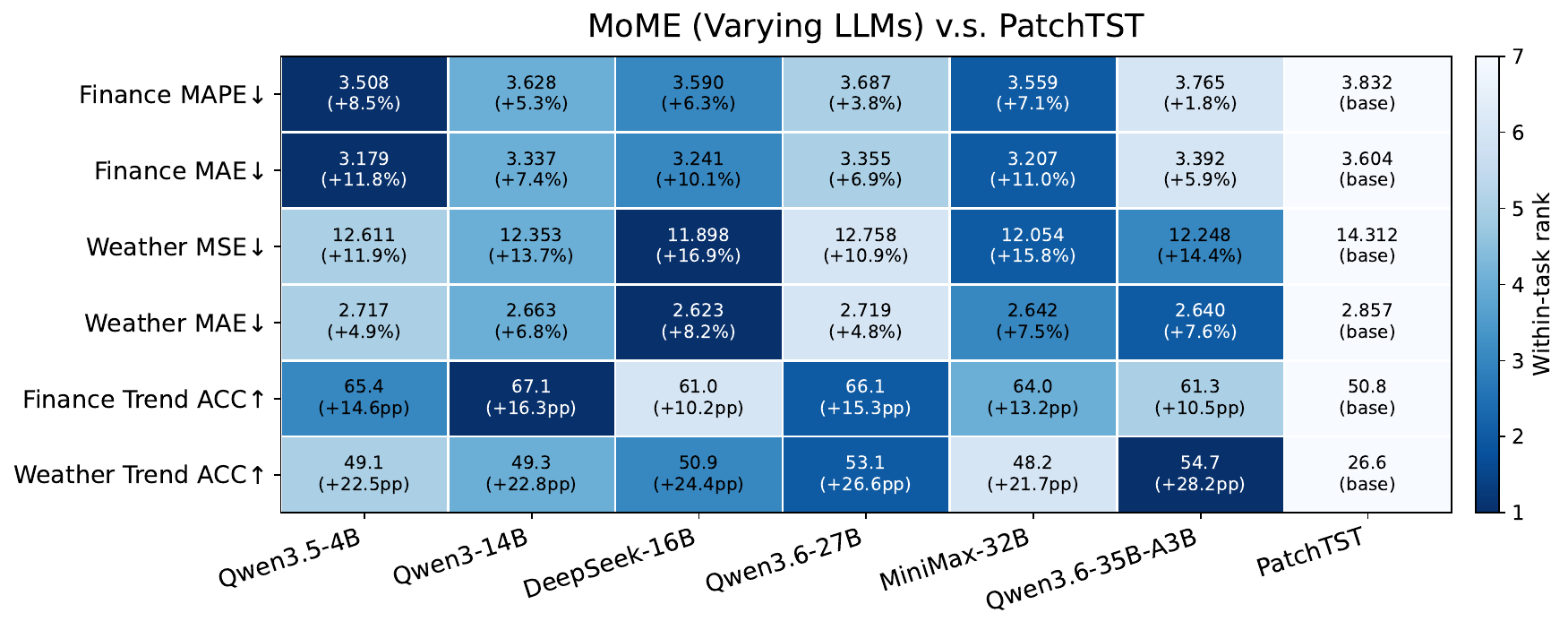}
    \caption{Comparison between MoME with varying LLM backbones and the PatchTST baseline. Each cell reports the model performance and its improvement over PatchTST. Detailed metric numbers for generating those figures are provided in Appendix~\ref{app:MoME_additional_tables}.}
    \label{fig:vary_llm}
\end{figure}

\paragraph{Hyperparameters.}
We conduct hyperparameter analysis on the critical design choices in our framework, including the number of activated experts $K$, the total number of experts $E$, and the number of context tokens. The results are summarized in Figure~\ref{fig:hyperparameter_fig} (b), with additional details in Appendix~\ref{app:MoME_additional_tables}. We observe that sparse expert activation often achieves stronger performance than denser alternatives. This empirical trend is also qualitatively consistent with the selective-computation motivation underlying Theorem~\ref{theorem:routing}. Since time series often contain noise, not all expert contributions are equally useful for every input, and selectively truncating a subset of them may help suppress less relevant computations. In addition, a moderate number of context tokens generally performs best, suggesting a trade-off between retaining informative semantic cues and avoiding unnecessary redundancy.

\paragraph{Different LLMs.}
We further evaluate MoME with different LLM backbones to assess the generalizability of the proposed framework. By conducting experiments on MT-Bench, we find the following two results shown in Figure~\ref{fig:vary_llm}. First, across different LLM backbones, MoME consistently outperforms the PatchTST baseline across all evaluated tasks. For forecasting, the improvements are relatively stable across models, while the gains are more pronounced for trend prediction. This suggests that the semantic information introduced through the LLM is particularly beneficial for high-level trend analysis beyond purely numerical forecasting. Second, we observe that the model performance does not scale monotonically with model size. The smallest backbone, Qwen3.5-4B, performs best on Finance forecasting, while mid-sized models such as DeepSeek-16B and Qwen3-14B achieve the best results on Weather forecasting and Finance trend prediction, respectively. The largest model, Qwen3.6-35B-A3B, performs best on Weather trend prediction. Overall, these results indicate that MoME is robust to the choice of LLM backbone and that its effectiveness is not simply driven by increasing model scale.

\subsection{Comparison with Token-Fusion Strategies}
\label{sec:fusion_compare_main}
We further compare our model with a series of token-fusion variants. To ensure a fair comparison, we only change the way of injecting multi-modal information while keeping other settings identical. Specifically, we first disable expert modulation to obtain a uni-modal variant, denoted as MoME ($w/o$\, EM). We then use this model as the backbone and attach different token-fusion modules, including early and late fusion strategies illustrated in Figure~\ref{fig:hyperparameter_fig} (a), to integrate textual information with time series input at the token level (implementation details in Appendix~\ref{app:fusion-compare}). We compare these models from multiple perspectives, including predictive performance, memory consumption, and inference speed. We find that the proposed expert-modulation method  achieves better performance and maintains favorable efficiency compared to token-fusion variants. 

\begin{figure}
    \centering
    \includegraphics[width=\linewidth]{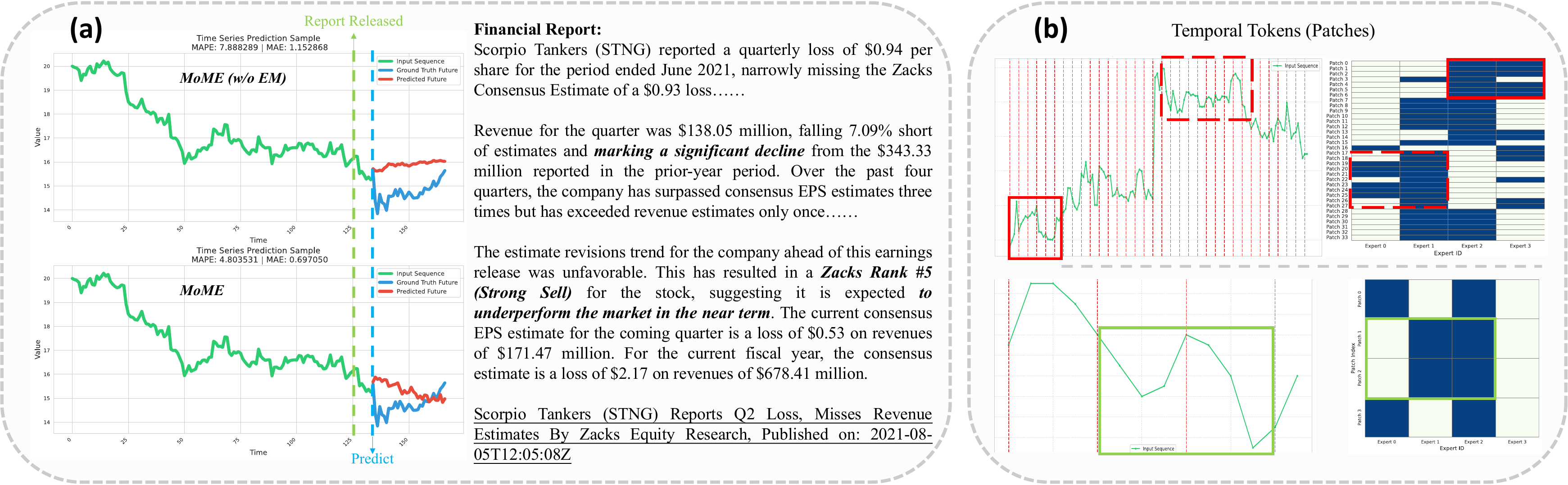}
    \caption{\textbf{(a)} A case study comparison between MoME with \textit{expert modulation} (EM) and its uni-modal variant ($w/o$ EM). \textbf{(b)} Routing patterns when activating EM.}
    \label{fig:main_case}
\end{figure}

\subsection{Routing Patterns}
Recent studies~\citep{chen2022towards, sun2024learning, guo2025advancing} have shown that MoE models tend to route similar tokens to similar subsets of experts, leading to expert specialization. We observe that MoME preserves this property when integrating multi-modal information. As shown in Figure~\ref{fig:main_case} (b, upper), temporal tokens with similar magnitudes are often routed to similar sets of experts. Beyond magnitude, we also find that the morphological patterns of the time series play a decisive role in expert selection (Figure~\ref{fig:main_case} (b, lower)). More examples are provided in Appendix~\ref{app:expert-select}.

This specialization behavior is a key factor underlying the effectiveness of MoE, as it enables different experts to focus on distinct regions of the input space and capture heterogeneous patterns more effectively. The preservation of such structure in MoME suggests that expert modulation does not disrupt but rather complements the intrinsic advantages of MoE.

\subsection{The Role of Multi-Modality}
\label{sec:question2}
To investigate how textual information influences the time series model, we conduct a case study on stock price prediction (Figure~\ref{fig:main_case} (a)). The historical price exhibits an overall downward trend, with a noticeable slowdown in the most recent period. As a result, the uni-modal variant, MoME ($w/o$ EM), extrapolates the recent pattern and predicts that the future price will remain relatively stable. In contrast, recent financial reports provide strong signals (multiple adverse market factors) indicating a potential further decline in the price. By incorporating these external textual signals, the full MoME model can anticipate the impending downward trend more accurately. This results in a substantial accuracy gain, reducing MAPE by nearly 50\% relative to uni-modal prediction. Importantly, in the examined cases where the external text is \textbf{misleading} or \textbf{inconsistent with subsequent developments}, MoME still produces reasonable predictions without substantial degradation relative to its uni-modal variant. These examples suggest that the model can remain relatively stable under imperfect textual signals. Additional case studies are provided in Appendix~\ref{app:case study}.


\section{Conclusion and Future Work}
\label{sec:conclusion}
We propose MoME, an MoE-based perspective for multi-modal integration that shifts the view of cross-modal interaction from conventional token-fusion to expert-modulation. Across a wide range of forecasting and trend prediction tasks, MoME improves upon strong uni-modal and multi-modal baselines, demonstrating that expert-modulation is an effective mechanism for integrating modality information. 

A key limitation of the current design is its restriction to one-way modulation, where it only allows LLM-derived textual signals to modulate time series experts; extending this to the reverse direction, allowing temporal signals to also influence LLM experts, may enable more advanced capabilities such as time series reasoning and question answering. In addition, while our study focuses on time series and text modalities, the expert modulation principle is not restricted to this setting and can be naturally generalized to scenarios involving more than two modalities by introducing additional modulation pathways and jointly conditioning expert behaviors. We believe these directions point to a broader role of expert modulation as a building block for future multi-modal learning systems.



\bibliography{tmlr}
\bibliographystyle{tmlr}


\appendix

\section{Detailed Theoretical Analysis}
\label{app:theoretical}

\subsection{Proof of Lemma~\ref{lemma:decompositoin}}
\label{app:theoretical1}

Consider an MLP layer of the following form:
\[
f(\mathbf{x})
=
\mathbf{W}^{(d)}
\left(
\sigma(\mathbf{W}^{(u)}\mathbf{x})
\odot
(\mathbf{W}^{(g)}\mathbf{x})
\right),
\]
where
$\mathbf{W}^{(u)},\mathbf{W}^{(g)}\in\mathbb{R}^{m\times d}$,
$\mathbf{W}^{(d)}\in\mathbb{R}^{r\times m}$, and $\mathbf{x}\in\mathbb{R}^{d}$. Here, $\sigma(\cdot)$ is an element-wise nonlinear activation, and $\odot$ denotes element-wise multiplication.

Let the hidden dimension be $m=E \times C$, and partition the $m$ hidden units into $E$ disjoint groups of size $C$. For each group $i=1,\ldots,E$, denote the corresponding submatrices by
\[
\mathbf{W}^{(u)}_i,\mathbf{W}^{(g)}_i
\in\mathbb{R}^{C\times d},
\qquad
\mathbf{W}^{(d)}_i
\in\mathbb{R}^{r\times C}.
\]

Define
\[
\mathbf{z}(\mathbf{x})
:=
\sigma(\mathbf{W}^{(u)}\mathbf{x})
\odot
(\mathbf{W}^{(g)}\mathbf{x}).
\]
Since both $\sigma(\cdot)$ and $\odot$ operate element-wise, $\mathbf{z}(\mathbf{x})$ can be partitioned according to the same $E$ groups:
\[
\mathbf{z}(\mathbf{x})
=
\begin{bmatrix}
\mathbf{z}_1(\mathbf{x})\\
\vdots\\
\mathbf{z}_E(\mathbf{x})
\end{bmatrix}
\in\mathbb{R}^{m},
\]
where
\[
\mathbf{z}_i(\mathbf{x})
=
\sigma\!\left(\mathbf{W}^{(u)}_i\mathbf{x}\right)
\odot
\left(\mathbf{W}^{(g)}_i\mathbf{x}\right)
\in\mathbb{R}^{C}.
\]

Similarly, partition $\mathbf{W}^{(d)}$ column-wise into matching blocks:
\[
\mathbf{W}^{(d)}
=
\begin{bmatrix}
\mathbf{W}^{(d)}_1 &
\cdots &
\mathbf{W}^{(d)}_E
\end{bmatrix}.
\]
Then,
\begin{align*}
f(\mathbf{x})
&=
\mathbf{W}^{(d)}\mathbf{z}(\mathbf{x}) =
\sum_{i=1}^{E}
\mathbf{W}^{(d)}_i\mathbf{z}_i(\mathbf{x})\\
&=
\sum_{i=1}^{E}
\mathbf{W}^{(d)}_i
\left(
\sigma\!\left(\mathbf{W}^{(u)}_i\mathbf{x}\right)
\odot
\left(\mathbf{W}^{(g)}_i\mathbf{x}\right)
\right).
\end{align*}

Defining
\[
f_i(\mathbf{x})
:=
\mathbf{W}^{(d)}_i
\left(
\sigma\!\left(\mathbf{W}^{(u)}_i\mathbf{x}\right)
\odot
\left(\mathbf{W}^{(g)}_i\mathbf{x}\right)
\right),
\]
we obtain
\[
f(\mathbf{x})
=
\sum_{i=1}^{E}f_i(\mathbf{x}),
\]
which completes the proof.

\subsection{Proof of Theorem~\ref{theorem:routing}}
\label{app:theoretical2}
A fully-activated MoE with $E$ experts can be expressed as
\[
    \mathrm{MoE}_{D}(\mathbf{x}) = \sum_{i=1}^{E}  g_i(\mathbf{x})f_i(\mathbf{x}).
\]
A sparse MoE under Top-K activation can be denoted as
\[
    \mathrm{MoE}_{S}(\mathbf{x}) = \sum_{i \in \mathcal{A}}g_i(\mathbf{x})f_i(\mathbf{x})=\sum_{i=1}^{E} \lambda_ig_i(\mathbf{x})f_i(\mathbf{x}),
\]
where $\sum_i\lambda_i = K, \lambda_i = \{0,1\}$ denotes the binary mask for expert selection and $\mathcal{A}$ denotes the set of activated experts ($|\mathcal{A}|=K$), then the amount of truncated information is
\begin{align*}
    \mathcal{L}(\mathbf{x})
    &= \big\|\mathrm{MoE}_D(\mathbf{x}) - \mathrm{MoE}_S(\mathbf{x})\big\|^2 
    = \Bigl\|\sum_{i=1}^E g_i(\mathbf{x})f_i(\mathbf{x}) \;-\;
    \sum_{i \in \mathcal{A}} g_i(\mathbf{x})f_i(\mathbf{x})\Bigr\|^2\\
    &= \Big\|\sum_{i \notin \mathcal{A}} g_i(\mathbf{x}) f_i(\mathbf{x})\Big\|^2
\end{align*}
The above equation can be reorganized into vector notations. Let $\mathcal{I} := \{1,\dots,E\}\setminus\mathcal{A}$ be the index set of pruned experts, so that $|\mathcal{I}| = E-K$. Then, we define
\[
    \mathbf{g}(\mathbf{x}) := (g_i(\mathbf{x}))_{i\in\mathcal{I}} \in \mathbb{R}^{E-K} \quad\text{and}\quad
    \mathbf{f}(\mathbf{x}) := [\,f_i(\mathbf{x})\,]_{i\in\mathcal{I}} \in \mathbb{R}^{d\times(E-K)},
\]
where $d$ is the dimension of the latent space. Let us denote
\[
\mathbf{F}(\mathbf{x})  = \mathbf{f}^\top(\mathbf{x}) \mathbf{f}(\mathbf{x}) \in \mathbb{R}^{(E-K) \times (E-K)},
\]
where $\mathbf{F}_{ij}(\mathbf{x}) = \langle f_i(\mathbf{x}), f_j(\mathbf{x})\rangle$, be the corresponding gram matrix. Then, we have
\begin{align*}
    \mathcal{L}(\mathbf{x}) &= \|\mathbf{f}(\mathbf{x})\mathbf{g}(\mathbf{x}) \|^2 = \langle \mathbf{f}(\mathbf{x})\mathbf{g}(\mathbf{x}), \mathbf{f}(\mathbf{x})\mathbf{g}(\mathbf{x})\rangle \\
    &= \mathbf{g}^\top(\mathbf{x})\mathbf{f}^\top(\mathbf{x})\mathbf{f}(\mathbf{x})\mathbf{g}(\mathbf{x}) = \mathbf{g}^\top(\mathbf{x})\mathbf{F}(\mathbf{x})\mathbf{g}(\mathbf{x}).
\end{align*}
Recall that a Rayleigh quotient is defined as
\[
    R(\mathbf{F}, \mathbf{g}) = \frac{\mathbf{g}^\top\mathbf{F}\mathbf{g}}{\mathbf{g}^\top\mathbf{g}}.
\]
Since $\mathbf{F}(\mathbf{x})$ is symmetric positive semi-definite, the Rayleigh quotient implies that
\begin{equation}
    \mathbf{g}^\top(\mathbf{x})\mathbf{F}(\mathbf{x})\mathbf{g}(\mathbf{x}) \leq \gamma_{max}(\mathbf{F}) \|\mathbf{g}(\mathbf{x})\|^2,
    \label{eq:quadratic_ieq}
\end{equation}
where $\gamma_{max}$ stands for the largest eigenvalue of the matrix $\mathbf{F}(\mathbf{x})$. By the Gershgorin circle theorem, for every eigenvalue $\gamma$ of $\mathbf{F}(x)$, we have the inequality:
\[
|\gamma - \mathbf{F}_{ii}| \leq \sum_{j \neq i}|\mathbf{F}_{ij}| \leq (E-K-1)\mu
\]
The maximum eigenvalue also suffices for the inequality. 
\[
\Rightarrow |\gamma_{max} - \mathbf{F}_{ii}| \leq (E-K-1)\mu.
\]
Using this bounding and the initial assumption ($\|f_i(\mathbf{x})\| \le B, \,\text{for}\, \forall i$), we have
\begin{align*}
    |\gamma_{max} | &\leq | \mathbf{F}_{ii}| + (E-K-1)\mu \\
    |\gamma_{max}| &\leq B^2+(E-K-1)\mu.
\end{align*}
Substituting this inequality into Equation (\ref{eq:quadratic_ieq}), we obtain
\[
    \mathcal{L}(\mathbf{x}) \leq [B^2+(E-K-1)\mu]\sum_{i \notin \mathcal{A}}g_i(\mathbf{x})^2,
\]
which completes the proof.

\subsection{Proof of Theorem~\ref{theorem:modulation}}
\label{app:theoretical3}

We first prove the result for the fully-activated MoE, where all experts participate in the aggregation and the routing coefficients are normalized. For sparse Top-$K$ activated MoE, the same proof applies by replacing the full expert set
\(\{1,\ldots,E\}\) with the active expert set \(\mathcal{A}\), provided that the selected routing coefficients are re-normalized (i.e., $\sum_{i\in\mathcal{A}} g_i(\mathbf{x}) = 1$), which is also commonly used in many MoE frameworks.

1. For Inclusion $\mathcal{F}_{\mathrm{fea}}\subseteq\mathcal{F}_{\mathrm{exp}}$.  

Take any $f_{\mathrm{fea}}\in\mathcal{F}_{\mathrm{fea}}$, i.e. $f_{\mathrm{fea}} = \gamma(\mathbf{z})F(\mathbf{x}) + \beta(\mathbf{z})$. In the expert modulation scheme, set $\gamma_i(\mathbf{z}) \equiv \gamma(\mathbf{z})$ and $\beta_i(\mathbf{z}) \equiv \beta(\mathbf{z})$ for all $i$. Then
\begin{align*}
    f_{\mathrm{exp}}(\mathbf{x}\mid\mathbf{z})&=\sum_{i=1}^{E}g_i(\mathbf{x})\left(\gamma_i(\mathbf{z})f_i(\mathbf{x})+\beta_i(\mathbf{z})\right)=\sum_{i=1}^{E}g_i(\mathbf{x})\left(\gamma(\mathbf{z})f_i(\mathbf{x})+\beta(\mathbf{z})\right) \\
    &=\gamma(\mathbf{z})\sum_{i=1}^{E}g_i(\mathbf{x})f_i(\mathbf{x})+\beta(\mathbf{z})\sum_{i=1}^{E}g_i(\mathbf{x}) = \gamma(\mathbf{z})F(\mathbf{x})+\beta(\mathbf{z})\\
    &=f_{\mathrm{fea}}(\mathbf{x}\mid\mathbf{z}).
\end{align*}
Hence, every feature‑modulated function can be achieved by expert modulation.

2. For Strictness $\mathcal{F}_{\mathrm{fea}}\neq\mathcal{F}_{\mathrm{exp}}$.

We next show that
$\mathcal{F}_{\mathrm{fea}}\neq\mathcal{F}_{\mathrm{exp}}$.
Let $h_i(\mathbf{x}):=g_i(\mathbf{x})f_i(\mathbf{x})$ denote the routed output of expert $i$.

Consider an expert-modulated function with
\[
\gamma_1(\mathbf{z})
\not\equiv
\gamma_2(\mathbf{z}),
\quad
\boldsymbol{\beta}_i(\mathbf{z})
\equiv
\mathbf{0},
\quad \forall i.
\]
Its output is
\[
f_{\mathrm{exp}}(\mathbf{x}\mid\mathbf{z})
=
\sum_{i=1}^{E}
\gamma_i(\mathbf{z})h_i(\mathbf{x}).
\]

Suppose for contradiction that
$f_{\mathrm{exp}}\in\mathcal{F}_{\mathrm{fea}}$.
Then there must exist a global scale $\gamma(\mathbf{z})$ and shift
$\boldsymbol{\beta}(\mathbf{z})$ such that
\[
\sum_{i=1}^{E}
\gamma_i(\mathbf{z})h_i(\mathbf{x})
=
\gamma(\mathbf{z})
\sum_{i=1}^{E}
h_i(\mathbf{x})
+
\boldsymbol{\beta}(\mathbf{z}).
\]
Rearranging gives
\[
\sum_{i=1}^{E}
\left(
\gamma_i(\mathbf{z})
-
\gamma(\mathbf{z})
\right)
h_i(\mathbf{x})
=
\boldsymbol{\beta}(\mathbf{z}).
\]

For any fixed $\mathbf{z}$, the right-hand side is constant with respect to $\mathbf{x}$. By the non-degeneracy assumption, no non-trivial linear combination of the routed expert outputs can be constant in $\mathbf{x}$. Therefore,
\[
\gamma_i(\mathbf{z})
-
\gamma(\mathbf{z})
=
0,
\quad \forall i,
\,\,\text{and}\,\,
\boldsymbol{\beta}(\mathbf{z})
=
\mathbf{0}.
\]
This implies $\gamma_i(\mathbf{z}) = \gamma(\mathbf{z})$, for $\forall i$, which contradicts the construction $\gamma_1(\mathbf{z})\not\equiv\gamma_2(\mathbf{z})$. Hence, there exist functions that are representable by expert modulation but cannot be represented by feature modulation. 

Combined with
$\mathcal{F}_{\mathrm{fea}}\subseteq\mathcal{F}_{\mathrm{exp}}$,
we conclude that $\mathcal{F}_{\mathrm{fea}} \subsetneq \mathcal{F}_{\mathrm{exp}}$.

\section{Selective Experts Computation for Uni-Modal Time Series Forecasting}
\label{app:ts-moe}

Theorem~\ref{theorem:routing} characterizes sparse MoE routing as a selective truncation of expert contributions. The theorem itself does not imply that sparse activation necessarily improves predictive performance; instead, it formalizes how Top-$K$ routing selectively retains a subset of expert computations. In this section, we empirically examine whether such selective expert computation is useful for time series forecasting. Our goals are: 

\textbf{(i)} to complement Theorem~\ref{theorem:routing} by comparing dense models and their sparse MoE variants in multiple time series forecasting tasks;

\textbf{(ii)} to investigate whether such sparse MoE can be incorporated into different time series model architectures and provide empirical benefits;

\textbf{(iii)} to examine whether the learned routing behavior exhibits meaningful expert specialization.

Together, these experiments provide empirical motivation for utilizing expert-wise computation in our multi-modal integration framework.

\subsection{Unified MoE Formulation}
Most existing MoE-based time series models introduce MoE by replacing the feed-forward layers of Transformer architectures with expert blocks~\citep{liu2024moirai, shi2024time}. Here, we consider a more general view: token-wise transformations in different time series architectures can be naturally generalized to an MoE form.

Time series models contain various token-wise mappings, including linear
projections, MLP blocks, and feed-forward layers, such as
\[
f(\mathbf{x}): \mathbb{R}^d \rightarrow \mathbb{R}^d,
\qquad 
f(\mathbf{x}): \mathbb{R}^L \rightarrow \mathbb{R}^{L'},
\]
where $d$, $L$, $L'$ denote the hidden dimensions and temporal window
lengths. We generalize such a mapping as
\[
f_{\mathrm{MoE}}(\mathbf{x}) = \sum_{i=1}^{E} \lambda_i(\mathbf{x})\, g_i(\mathbf{x})\, f_i(\mathbf{x})
\]
where $f_i(\cdot)$ denotes the $i$-th expert,
$g_i(\mathbf{x})$ is its routing coefficient, and $\lambda_i(\mathbf{x})\in\{0,1\}$ is the Top-$K$ selection mask. This formulation exposes an explicit set of expert-wise transformations and allows the model to selectively activate different computations for different inputs. We use this formulation to study sparse expert computation across several representative time series architectures.



\subsection{Backbone Integration}
Following the formulation above, we integrate MoE into several representative time series backbones spanning different architectural families, including linear models (DLinear~\citep{zeng2023transformers}), MLP-based models (TSMixer~\citep{chen2023tsmixer}), and transformer-based models (iTransformer~\citep{liu2023itransformer}). Specifically, for DLinear, the per-channel linear mappings in both the seasonal and trend components are replaced by linear-expert MoE; For TSMixer, the time-mixing MLP is replaced by MLP-expert MoE; For iTransformer, the feed-forward layers in the inverted transformer are similarly replaced by MLP-expert MoE. These different integrations allow us to examine how MoE interacts with different computational structures.

\begin{table}[]
    \centering
    \caption{Performance comparison of MoE integration across different models and datasets.}
    \resizebox{\linewidth}{!}{
    \begin{tabular}{llcccccccccccc|c}
\toprule
\multirow{2}{*}{Dataset} & \multirow{2}{*}{$h$} &
\multicolumn{4}{c}{\textbf{TSMixer}} & 
\multicolumn{4}{c}{\textbf{DLinear}} &
\multicolumn{4}{c}{\textbf{iTransformer}} &
\multirow{2}{*}{\textbf{Avg Imp (\%)}} \\
\cmidrule(lr){3-6} \cmidrule(lr){7-10} \cmidrule(lr){11-14}
& & \multicolumn{2}{c}{Original} & \multicolumn{2}{c}{+ MoE}
& \multicolumn{2}{c}{Original} & \multicolumn{2}{c}{+ MoE}
& \multicolumn{2}{c}{Original} & \multicolumn{2}{c}{+ MoE} & \\
\cmidrule(lr){3-4}\cmidrule(lr){5-6}
\cmidrule(lr){7-8}\cmidrule(lr){9-10}
\cmidrule(lr){11-12}\cmidrule(lr){13-14}
& & MSE & MAE & MSE & MAE & MSE & MAE & MSE & MAE & MSE & MAE & MSE & MAE & \\
\midrule

\multirow{4}{*}{ETTm1} 
& 96  & 0.294 & 0.345 & \textbf{0.288} & \textbf{0.343} & \textbf{0.286} & \textbf{0.335} & 0.304 & 0.348 & 0.320 & 0.368 & \textbf{0.318} & \textbf{0.364} & -0.95 \\
& 192 & \textbf{0.327} & \textbf{0.359} & 0.331 & 0.369 & \textbf{0.327} & \textbf{0.359} & 0.334 & 0.367 & 0.361 & 0.391 & \textbf{0.357} & \textbf{0.389} & -1.12 \\
& 336 & \textbf{0.374} & 0.394 & \textbf{0.374} & 0.398 & \textbf{0.367} & \textbf{0.381} & \textbf{0.367} & \textbf{0.381} & \textbf{0.393} & \textbf{0.411} & 0.396 & \textbf{0.411} & -0.29 \\
& 720 & \textbf{0.429} & \textbf{0.424} & \textbf{0.429} & 0.429 & \textbf{0.428} & \textbf{0.417} & 0.439 & 0.434 & 0.448 & 0.439 & \textbf{0.442} & \textbf{0.438} & -1.04 \\
\midrule

\multirow{4}{*}{ETTm2}
& 96  & \textbf{0.168} & \textbf{0.255} & 0.170 & \textbf{0.255} & 0.179 & 0.274 & \textbf{0.177} & \textbf{0.265} & 0.181 & 0.273 & \textbf{0.180} & \textbf{0.272} & 0.67 \\
& 192 & 0.234 & 0.300 & \textbf{0.228} & \textbf{0.296} & \textbf{0.241} & 0.317 & \textbf{0.241} & \textbf{0.313} & 0.249 & 0.317 & \textbf{0.247} & \textbf{0.314} & 1.15 \\
& 336 & \textbf{0.278} & \textbf{0.332} & 0.280 & \textbf{0.332} & 0.399 & 0.417 & \textbf{0.297} & \textbf{0.349} & 0.299 & 0.347 & \textbf{0.294} & \textbf{0.344} & 7.28 \\
& 720 & \textbf{0.398} & 0.406 & \textbf{0.398} & \textbf{0.404} & 0.738 & 0.562 & \textbf{0.430} & \textbf{0.438} & 0.376 & 0.397 & \textbf{0.369} & \textbf{0.392} & 12.24 \\
\midrule

\multirow{4}{*}{ETTh1}
& 96  & 0.392 & 0.411 & \textbf{0.383} & \textbf{0.407} & 0.379 & 0.399 & \textbf{0.375} & \textbf{0.396} & 0.402 & 0.418 & \textbf{0.393} & \textbf{0.414} & 1.38 \\
& 192 & 0.419 & 0.430 & \textbf{0.415} & \textbf{0.427} & 0.422 & 0.429 & \textbf{0.411} & \textbf{0.421} & 0.441 & 0.442 & \textbf{0.434} & \textbf{0.441} & 1.32 \\
& 336 & 0.442 & \textbf{0.441} & \textbf{0.438} & 0.444 & 0.451 & \textbf{0.448} & \textbf{0.446} & \textbf{0.448} & 0.453 & 0.454 & \textbf{0.443} & \textbf{0.448} & 0.81 \\
& 720 & 0.508 & 0.496 & \textbf{0.487} & \textbf{0.488} & 0.490 & 0.489 & \textbf{0.485} & 0.499 & 0.472 & 0.484 & \textbf{0.468} & \textbf{0.479} & 1.11 \\
\midrule

\multirow{4}{*}{ETTh2}
& 96  & \textbf{0.313} & 0.366 & \textbf{0.313} & \textbf{0.364} & 0.370 & 0.393 & 0.302 & \textbf{0.360} & 0.319 & 0.372 & \textbf{0.310} & \textbf{0.367} & 5.25 \\
& 192 & 0.360 & 0.395 & \textbf{0.358} & \textbf{0.394} & 0.495 & 0.466 & \textbf{0.391} & \textbf{0.417} & 0.375 & 0.413 & \textbf{0.374} & \textbf{0.411} & 5.51 \\
& 336 & 0.390 & 0.417 & \textbf{0.376} & \textbf{0.412} & 0.595 & 0.524 & \textbf{0.410} & \textbf{0.436} & \textbf{0.398} & \textbf{0.428} & \textbf{0.398} & \textbf{0.428} & 8.78 \\
& 720 & \textbf{0.414} & \textbf{0.437} & 0.417 & 0.438 & 0.888 & 0.656 & \textbf{0.833} & \textbf{0.649} & 0.426 & 0.452 & \textbf{0.421} & \textbf{0.448} & 1.39 \\
\midrule

\multirow{4}{*}{EXR}
& 96  & 0.099 & 0.228 & \textbf{0.095} & \textbf{0.222} & 0.089 & 0.218 & \textbf{0.081} & \textbf{0.205} & 0.110 & 0.240 & \textbf{0.106} & \textbf{0.237} & 4.42 \\
& 192 & \textbf{0.196} & \textbf{0.320} & \textbf{0.196} & 0.321 & \textbf{0.172} & 0.313 & \textbf{0.172} & \textbf{0.309} & 0.236 & 0.354 & \textbf{0.227} & \textbf{0.348} & 1.08 \\
& 336 & 0.369 & 0.446 & \textbf{0.352} & 0.439 & 0.334 & 0.444 & \textbf{0.332} & \textbf{0.434} & 0.405 & 0.477 & \textbf{0.401} & \textbf{0.471} & 1.88 \\
& 720 & 0.965 & 0.740 & \textbf{0.881} & \textbf{0.716} & 0.879 & 0.728 & \textbf{0.767} & \textbf{0.659} & 0.849 & 0.698 & \textbf{0.802} & \textbf{0.673} & 7.21 \\
\midrule

\multirow{4}{*}{ECL}
& 96  & 0.139 & 0.238 & \textbf{0.138} & \textbf{0.235} & 0.136 & \textbf{0.233} & \textbf{0.135} & \textbf{0.233} & 0.164 & 0.271 & \textbf{0.160} & \textbf{0.264} & 1.29 \\
& 192 & 0.161 & \textbf{0.257} & \textbf{0.159} & \textbf{0.257} & \textbf{0.151} & 0.248 & \textbf{0.151} & \textbf{0.246} & 0.185 & 0.289 & \textbf{0.179} & \textbf{0.281} & 1.34 \\
& 336 & 0.171 & 0.271 & \textbf{0.167} & \textbf{0.267} & \textbf{0.166} & 0.265 & 0.168 & \textbf{0.265} & 0.201 & 0.304 & \textbf{0.194} & \textbf{0.294} & 1.56 \\
& 720 & 0.200 & 0.294 & \textbf{0.192} & \textbf{0.287} & \textbf{0.203} & 0.300 & \textbf{0.203} & \textbf{0.298} & 0.239 & 0.333 & \textbf{0.228} & \textbf{0.321} & 2.54 \\
\midrule

\multirow{4}{*}{TRF}
& 96  & 0.386 & 0.273 & \textbf{0.378} & \textbf{0.273} & 0.427 & 0.302 & \textbf{0.401} & \textbf{0.278} & 0.374 & 0.276 & \textbf{0.371} & \textbf{0.271} & 3.73 \\
& 192 & 0.414 & 0.289 & \textbf{0.399} & \textbf{0.276} & 0.441 & 0.308 & \textbf{0.422} & \textbf{0.286} & 0.400 & 0.285 & \textbf{0.379} & \textbf{0.283} & 3.50 \\
& 336 & 0.431 & 0.302 & \textbf{0.410} & \textbf{0.283} & 0.455 & 0.317 & \textbf{0.439} & \textbf{0.294} & 0.415 & 0.294 & \textbf{0.406} & \textbf{0.287} & 4.41 \\
& 720 & 0.463 & 0.328 & \textbf{0.450} & \textbf{0.312} & 0.487 & 0.335 & \textbf{0.465} & \textbf{0.313} & 0.452 & 0.321 & \textbf{0.448} & \textbf{0.311} & 3.80 \\
\midrule

\multicolumn{15}{r}{\textbf{Overall Avg Imp (\%): 2.83}} \\  
\bottomrule
\end{tabular}
    }
    \label{tab:encoder_classic}
\end{table}
\begin{table}[]
    \centering
    \caption{Zero-shot performance on ETT datasets. The best results between Original and the MoE variant for each model are in bold.}
    \resizebox{\linewidth}{!}{
    \begin{tabular}{llcccccccccccccccc}
\toprule
\multirow{2}{*}{Dataset} & \multirow{2}{*}{$h$} &
\multicolumn{2}{c}{\textbf{TSMixer}} &
\multicolumn{2}{c}{\textbf{TSMixer (MoE)}} &
\multicolumn{2}{c}{\textbf{DLinear}} &
\multicolumn{2}{c}{\textbf{DLinear (MoE)}} &
\multicolumn{2}{c}{\textbf{iTransformer}} &
\multicolumn{2}{c}{\textbf{iTransformer (MoE)}} &
\multicolumn{2}{c}{\textbf{Imp (\%)}} \\
\cmidrule(lr){3-4}\cmidrule(lr){5-6}
\cmidrule(lr){7-8}\cmidrule(lr){9-10}
\cmidrule(lr){11-12}\cmidrule(lr){13-14}
\cmidrule(lr){15-16}
& & MSE & MAE & MSE & MAE & MSE & MAE & MSE & MAE & MSE & MAE & MSE & MAE & MSE & MAE \\
\midrule

\multirow{2}{*}{ETTh1 $\rightarrow$ ETTh2} 
& 96  & 0.304 & 0.358 & \textbf{0.302} & \textbf{0.357} & 0.310 & 0.371 & \textbf{0.292} & \textbf{0.355} & 0.288 & 0.353 & \textbf{0.284} & \textbf{0.351} & 2.617 & 1.720 \\
& 336 & 0.382 & 0.418 & \textbf{0.380} & \textbf{0.414} & 0.443 & 0.471 & \textbf{0.432} & \textbf{0.450} & 0.394 & 0.419 & \textbf{0.375} & \textbf{0.409} & 2.610 & 2.600 \\
\midrule

\multirow{2}{*}{ETTh2 $\rightarrow$ ETTh1}
& 96  & 0.519 & 0.483 & \textbf{0.513} & \textbf{0.483} & 0.459 & 0.450 & \textbf{0.424} & 0.430 & \textbf{0.617} & 0.542 & 0.619 & \textbf{0.541} & 2.819 & 1.543 \\
& 336 & 0.761 & 0.605 & \textbf{0.639} & \textbf{0.552} & 0.503 & 0.488 & \textbf{0.460} & \textbf{0.460} & 0.832 & 0.647 & \textbf{0.787} & \textbf{0.628} & 9.996 & 5.812 \\
\midrule

\multirow{2}{*}{ETTh2 $\rightarrow$ ETTm1}
& 96  & 0.724 & \textbf{0.656} & \textbf{0.719} & 0.661 & 0.737 & 0.657 & \textbf{0.715} & \textbf{0.647} & 1.017 & 0.772 & \textbf{0.991} & \textbf{0.762} & 2.077 & 0.685 \\
& 336 & 1.037 & 0.777 & \textbf{1.019} & \textbf{0.771} & 1.174 & 0.817 & \textbf{0.925} & \textbf{0.741} & 1.310 & 0.854 & \textbf{1.275} & \textbf{0.843} & 8.539 & 3.788 \\
\midrule

\multirow{2}{*}{ETTh1 $\rightarrow$ ETTm2}
& 96  & 0.889 & 0.672 & \textbf{0.872} & \textbf{0.667} & 0.990 & 0.703 & \textbf{0.928} & \textbf{0.685} & 1.022 & 0.717 & \textbf{1.015} & \textbf{0.716} & 2.954 & 1.148 \\
& 336 & 1.338 & 0.841 & \textbf{1.306} & \textbf{0.834} & 1.563 & 0.907 & \textbf{1.538} & \textbf{0.901} & 1.371 & 0.855 & \textbf{1.363} & \textbf{0.851} & 1.525 & 0.653 \\
\midrule

\multirow{2}{*}{ETTm1 $\rightarrow$ ETTm2}
& 96  & 0.192 & 0.275 & \textbf{0.182} & \textbf{0.164} & 0.186 & 0.278 & \textbf{0.183} & \textbf{0.275} & 0.209 & 0.285 & \textbf{0.203} & \textbf{0.282} & 3.230 & 14.166 \\
& 336 & 0.299 & 0.341 & \textbf{0.293} & \textbf{0.339} & \textbf{0.307} & \textbf{0.371} & 0.316 & 0.376 & 0.329 & 0.358 & \textbf{0.325} & \textbf{0.357} & 0.097 & -0.161 \\
\midrule

\multirow{2}{*}{ETTm2 $\rightarrow$ ETTm1}
& 96  & 0.530 & 0.464 & \textbf{0.487} & \textbf{0.448} & 0.384 & 0.400 & \textbf{0.373} & \textbf{0.390} & \textbf{0.521} & \textbf{0.474} & \textbf{0.521} & \textbf{0.474} & 3.659 & 1.982 \\
& 336 & 0.951 & 0.642 & \textbf{0.618} & \textbf{0.523} & 0.445 & 0.442 & \textbf{0.445} & \textbf{0.439} & \textbf{0.600} & \textbf{0.520} & 0.605 & 0.525 & 11.394 & 6.084 \\
\midrule

\multirow{1}{*}{\textbf{Avg Imp.}}
&  &  &  &  &  &  &  & & &  && & & 4.293 & 3.335 \\
\bottomrule
\end{tabular}
    }
    \label{tab:encoder_zero_shot}
\end{table}

\subsection{Results}
\label{app:uni_results}

\paragraph{Sparse MoE across Backbones.}
We first evaluate the sparse MoE variants across multiple time series benchmarks. Results are reported in Table~\ref{tab:encoder_classic}. Despite substantial differences among the three backbone architectures, incorporating sparse MoE generally improves forecasting performance over their original dense counterparts. The gains are particularly noticeable for some lightweight models such as DLinear, suggesting that selective expert decomposition can also benefit simple forecasting architectures.

We observe similar improvements under zero-shot forecasting (Table~\ref{tab:encoder_zero_shot}), indicating that the benefit of expert decomposition is not restricted to the training distribution. Overall, these results show that sparse expert computation can serve as a general architectural extension across different time series backbones.

\begin{figure}
    \centering
    \includegraphics[width=0.95\linewidth]{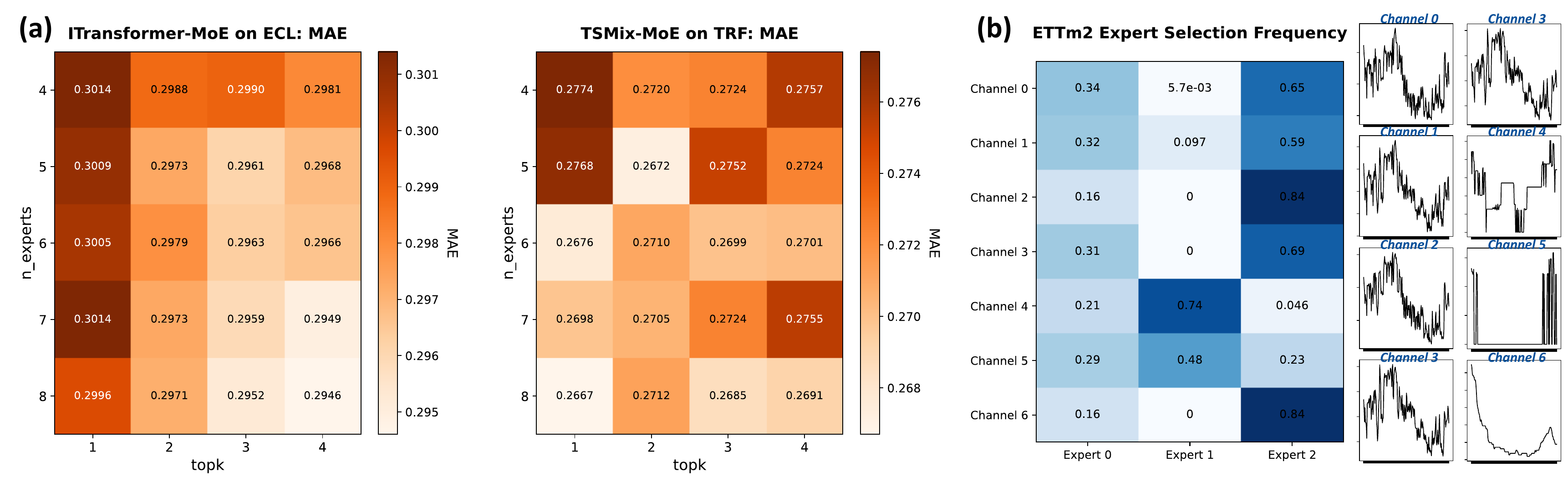}
    \caption{\textbf{(a)} Sparse expert activation could outperform dense activation for UMTSP models; \textbf{(b)} expert routing behavior on ETTm2 dataset, treating each channel as a token.}
    \label{fig:MoE_Uni_Classic}
\end{figure}

\paragraph{More Evidence on the Effect of Sparse Activation.}
We next isolate the role of expert sparsity by comparing different degrees of Top-$K$ activation. As shown in Figure~\ref{fig:MoE_Uni_Classic} (a), denser expert activation does not necessarily lead to better performance. These results complement the findings in Table~\ref{tab:encoder_classic} and Table~\ref{tab:encoder_zero_shot}, further suggesting that \textbf{selective expert computation can be beneficial for time series forecasting}. Together, the theoretical and empirical results suggest that explicitly controlling which expert computations participate in prediction provides a useful inductive bias for time series models.

\subsection{Expert Routing Behavior}
\label{app:uni_routing}
We also examine whether selective routing gives rise to meaningful expert specialization. We treat channels as tokens and visualize the routing patterns learned on ETTm2 in Figure~\ref{fig:MoE_Uni_Classic}(b). Channels 0--3 exhibit similar temporal dynamics, whereas Channels 4 and 5 show distinctly different patterns; Channel 6 resembles a smoother version of the first group. Correspondingly, Channels 4 and 5 are predominantly routed to Expert 1, while the remaining channels show a strong preference for Expert 2. Thus, channels with similar temporal patterns tend to activate similar subsets of experts.

This behavior suggests that the router learns to organize temporal inputs according to their underlying dynamics, giving rise to pattern-specific expert specialization. Such behavior is consistent with prior observations of specialization in MoE models in different domains~\citep{deepseekai2025deepseekv3technicalreport, guo2025advancing}. More importantly, for our framework, it indicates that the expert-wise decomposition exposed by MoE is meaningful rather than arbitrary: different temporal patterns are processed through different expert subsets. These observations serve as strong motivation for our multi-modal design, where auxiliary information is used to condition this expert-wise computation.


\subsection{Hyperparameter Searching}
Optuna~\citep{akiba2019optuna} to tune hyperparameters on the baseline models, and evaluate MoE variants under the same configurations. The search space includes learning rate $\sim \mathcal{U}(1\mathrm{e}{-5}, 5\mathrm{e}{-3})$, dropout $\sim \mathcal{U}(0.0, 0.6)$, model dimension $\in \{16, 32, 64, 128\}$, number of heads $\in \{2, 4, 8\}$, number of layers $\in [2, 6]$, patch length $\in \{8, 16, 32\}$, and stride $\in \{4, 8, 16\}$. We minimize validation MSE, apply \textit{MedianPruner} for early stopping, and run 30 trials per experiment.

\section{Dataset}
\label{app:data}

\subsection{General Discussion on MMTSP datasets}
\label{app:data0}

\paragraph{Time-Text Alignment Assumption.}
A key assumption in MMTSP is that the textual input should be relevant to the prediction target. We do not require strict timestamp-level alignment between every numerical observation and every textual report. Instead, we adopt a \textit{soft alignment} assumption: (i) \textbf{availability}, where the text can be obtained from the look-back observation window, and (ii) \textbf{relevance}, which indicates it provides information related to the prediction target. This assumption is both necessary and practically realistic, as informative textual signals must reflect recent observable conditions while remaining accessible prior to the prediction horizon.

\paragraph{Data Leakage Discussion.}
A common concern in MMTSP is whether textual inputs introduce data leakage. In our real-world experiments, we mitigate this issue by ensuring that all textual information is strictly retrieved before the prediction window. Under this condition, the text represents available contextual information rather than leaked future labels. While textual inputs may contain predictive cues about future outcomes, this is expected and does not constitute leakage, as such information would also be available in real-world decision-making scenarios. Therefore, these signals are legitimate forecasting inputs as long as they are observed prior to the prediction horizon.

\paragraph{Raw Text vs. Structured Indicators.}
Another possible strategy is to manually convert text into structured indicators. While such indicators can be compact and easier to learn from, their construction requires domain-specific expertise and design choices, such as what information to extract or how to quantify it. In addition, these manually designed indicators are difficult to generalize across domains.

Raw text provides more flexible and complementary information, but is also noisier than structured indicators. Our framework can be viewed as a data-driven compromise: rather than relying on manually designed indicators, MoME learns a trainable indicator directly from supervision (step 1 in Section~\ref{sec:mome}). This enables the model to extract useful signals from raw text while maintaining the flexibility to generalize across domains without task-specific redesign.

\begin{table}[]
    \centering
    \caption{Dataset used for MMTSP evaluations.}
    \resizebox{\linewidth}{!}{
    \begin{tabular}{llcccr}
\toprule
\textbf{Domain} & \textbf{Target} & \textbf{Dimension} & \textbf{Frequency} & \textbf{Number of Samples} & \textbf{Window per Sample} \\
\midrule
\textbf{MT-Bench (Real):}\\
MT-Weather (Short)         & Temperature               & 1   & 1 Hour & 1820   & 7 Days $\rightarrow$ 1 Day \\
MT-Weather (Long)          & Temperature             & 1   & 1 Hour  & 1841  & 14 Days $\rightarrow$ 3 Days \\
MT-Finance (Short)     & Stock Price  & 1  & 5 Minutes  & 916  & 7 Days $\rightarrow$ 1 Day \\
MT-Finance (Long)    & Stock Price & 1  & 1 Hour  & 1458  & 30 Days $\rightarrow$ 7 Days \\
\midrule
\textbf{Time-IMM (Real):}\\
GDELT (CAN) & Canada Event & 5 & 1 Day & 759 & 14 Days $\rightarrow$ 3 Days \\ 
EPA-Air & Temperature & 4 & 6 Hours & 1276 & 42 Hours $\rightarrow$ 9 Hours \\ 
ILINet & Infectious Status & 11 & 1 Week & 624 & 14 Weeks $\rightarrow$ 3 Weeks  \\ 
\midrule
\textbf{Time-MMD (Synthetic):}\\
Environment         & Air Quality               & 4   & 1 Day  & 1990 & 7 Days $\rightarrow$ 1 Day \\ 
Energy          & Gasoline Prices             & 9   & 1 Week & 718 & 14 Weeks $\rightarrow$ 3 Weeks \\ 
Health (US)     & Infectious Proportion & 11  & 1 Week  & 615 & 14 Weeks $\rightarrow$ 3 Weeks \\ 
Health (AFR)    & Infectious Proportion & 11  & 1 Week & 630 & 14 Weeks $\rightarrow$ 3 Weeks \\ 
Social Good     & Unemployment Rate             & 1   & 1 Month   & 436 & 14 Months $\rightarrow$ 3  Months \\ 
\bottomrule
\end{tabular}
    }
    \label{tab:data}
\end{table}
\begin{table}[]
    \centering
    \caption{Dataset used for UMTSP evaluations.}
    \resizebox{\linewidth}{!}{
    \begin{tabular}{lcccr}
\toprule
\textbf{Dataset} & \textbf{Target} & \textbf{Dimension} & \textbf{Total Timestamps} & \textbf{Frequency} \\
\midrule
ETTh1 / ETTh2   & Oil Temperature & 7     & 17{,}420  & 1 hour \\
ETTm1 / ETTm2   & Oil Temperature  & 7     & 69{,}680  & 15 min \\
Electricity     & Electricity Consumption & 321   & 26{,}304  & 1 hour \\
Traffic         & Road Occupancy Rate & 862   & 17{,}544  & 1 hour \\
\bottomrule
\end{tabular}
    }
    \label{tab:data_classic}
\end{table}
\begin{table}
    \centering
    \caption{Trend label binning for MT-Finance and MT-Weather data.}
    \resizebox{\textwidth}{!}{
    \begin{tabular}{llcc}
\toprule
\textbf{3-way Label} & \textbf{5-way Label} & \textbf{Financial Price} & \textbf{Weather Temperature} \\
\midrule
\multirow{2}{*}{Negative} 
  & Bearish & $< -4\%$ & Past: $< -0.25$ \quad Future: $< -1.5$ \\
  & Warning & $-4\% \sim -2\%$ & \\
\midrule
Neutral & Neutral & $-2\% \sim 2\%$ & Past: $-0.25 \sim 0.25$ \quad Future: $-1.5 \sim 1.5$ \\
\midrule
\multirow{2}{*}{Positive} 
  & Growth-Oriented & $2\% \sim 4\%$ & Past: $> 0.25$ \quad Future: $> 1.5$ \\
  & Bullish & $> 4\%$ & \\
\bottomrule
\end{tabular}
    }
    \label{tab:trend_labels}
\end{table}

\subsection{Real-World MMTSP Datasets}
\label{app:data1}

Our main MMTSP evaluation is conducted on real-world datasets. We use MT-Bench~\citep{chen2025mtbench} as the primary testbed, including Weather and Finance datasets. We also include complementary datasets (ILINet, EPA-Air, and GDELT) from Time-IMM~\citep{chang2026timeimm}. Datasets are detailed in Table~\ref{tab:data}.

\subsubsection{MT-Bench.}

\paragraph{Datasets.}
The finance dataset includes stock price time series paired with financial reports, which provides market-related information such as company-specific events or analyst opinions. The weather dataset includes temperature time series paired with weather-related reports, which provides more comprehensive meteorological information, such as the movement of the warm front.

\paragraph{Assignment of Trend Labels.}
To perform trend prediction, we convert continuous forecasting targets into discrete trend labels. For finance time series, we compute the percentage price change over the prediction window:
\[
    \Delta_{\text{finance}} = \frac{y_{\text{end}} - y_{\text{start}}}{y_{\text{start}}}.
\]
This value reflects whether the stock price increases, decreases, or remains stable.

For weather time series, we define trends differently for past and future windows. For past trend analysis, we downsample the input sequence into daily means and compute a linear slope:
\[
    \Delta_{\text{past}} = \mathrm{slope}(\{\bar{x}_i\}_{i=1}^{L}).
\]
For future trend prediction, since the prediction horizon is relatively short, we define the future trend as the mean temperature difference between the future window and the last input segment:
\[
    \Delta_{\text{future}} =
    \mathrm{Mean}(\{\bar{y}_i\}_{i=1}^{T})
    -
    \mathrm{Mean}(\{\bar{x}_j\}_{j=1}^{L_{\text{tail}}}).
\]
The resulting continuous trend values are discretized into categorical labels. All binning thresholds are summarized in Table~\ref{tab:trend_labels}.

\begin{wrapfigure}{r}{0.55\linewidth}
\begin{minipage}{\linewidth}
\refstepcounter{myalgorithm}
\label{alg:metrics}
\hrule
\centering
\textbf{Algorithm \themyalgorithm: ComputeMetrics$(x_{\mathrm{in}}, x_{\mathrm{out}})$}

\hrule
\raggedright
\textbf{Input:} Input and output windows $x_{\mathrm{in}}$, $x_{\mathrm{out}}$\\

\textbf{Output:} Directional and fluctuation metrics

\textbf{A. Near-term Narrative:}
Compare the mean of $x_{\mathrm{out}}$ with the tail of $x_{\mathrm{in}}$ to capture short-term level change.
\smallskip

\textbf{B. Severity:}
Estimate a scale parameter $\sigma$ from $x_{\mathrm{in}}$ using standard deviation, IQR, and MAD to ensure stability against outliers.
\smallskip

\textbf{C. Background Tendency:}
Fit a linear slope to $x_{\mathrm{in}}$ and normalize by $\sigma$ to obtain a trend score $z_{\mathrm{past}}$, which determines direction and strength.
\smallskip

\textbf{D. Past-Window Variability:}
Detrend $x_{\mathrm{in}}$ and compute residual volatility, smoothness ($R^2$), and normalized mean absolute difference.
\smallskip

\textbf{E. (Future) Pattern Cue:}
Fit a slope on $x_{\mathrm{out}}$, normalize by $\sigma$, and record the resulting direction and magnitude.

\textbf{Return:}
$\mathrm{Metrics}=\{\mathrm{A},\mathrm{B},\mathrm{C},\mathrm{D},\mathrm{E}\}$

\hrule
\end{minipage}
\end{wrapfigure}

\subsubsection{Time-IMM.}

\paragraph{Datasets.}
The ILINet dataset includes epidemiological time series with news reports, which provides information such as influenza immunization rates. EPA-Air includes temperature time series, paired with news reports such as environmental disaster warnings. GDELT includes social event time series (characterized by features such as event intensity and sentiment) paired with news reports, which provides information such as market signal propagation or stakeholder conflict dynamics.

\paragraph{Pre-Processing.}
Since the data provided by~\citep{chang2026timeimm} does not contain pre-aligned time-text pairs, we construct the pairing as follows. For each forecast sample, we select textual information satisfying $t_{\text{text\_release}} < t_{\text{prediction\_start}}$. This ensures that all textual inputs are observed before the prediction window. When multiple textual candidates are available, the most recent one is chosen to maximize temporal relevance.

\subsection{Semi-Synthetic MMTSP Dataset}
\label{app:data3}

In many real-world MMTSP datasets, textual information may be weakly aligned with the prediction target and may contain substantial noise. To \textbf{complement} our real-world evaluation, we construct a semi-synthetic benchmark based on Time-MMD~\citep{liu2024time}, where the textual modality is designed to contain more focused and complementary information about the underlying time series. Specifically, we use five datasets from Time-MMD, covering environmental (e.g., air quality), socio-economic (e.g., unemployment rates), and health (e.g., infectious disease) domains. Detailed dataset statistics are provided in Table~\ref{tab:data}.

The original textual inputs in Time-MMD often contain missing values and provide limited structured contextual information, which can make them weakly informative for MMTSP evaluation. To construct a more controlled setting, we generate synthetic reports from summary statistics of both the historical window $\mathbf{x}_{\mathrm{in}}$ and the future window $\mathbf{x}_{\mathrm{out}}$. The future window is used only to provide qualitative high-level cues for text generation, while the prompt explicitly prohibits exact future values, numerical forecasts, and direct target labels. We therefore treat Time-MMD as a \emph{target-informed semi-synthetic benchmark} for MMTSP evaluation. Its purpose is to provide more coherent and informative auxiliary text for controlled comparison of multi-modal integration mechanisms. Since all compared methods receive exactly the same synthetic textual inputs, this setting allows us to \textbf{examine how effectively different integration strategies utilize the same auxiliary information}. The corresponding experimental results are reported in Appendix~\ref{app:mome,time-mmd}. We use this benchmark only as a complementary analysis, while our main evaluation is conducted on real-world datasets. The synthesis is conducted by Algorithm~\ref{alg:metrics} and the LLM prompt~\ref{prompt_template}. We also present a comparison between the original and synthetic textual inputs~\ref{synthetic example}.

\begin{tcolorbox}[title={Example Prompt (Air-Quality Domain)}]
You are an environmental journalist. Write one concise English bulletin ($\sim180$ words) about \textbf{recently observed} air-quality conditions over the past \{\texttt{n\_days}\} days.

Guidance (not a forecast): 
\begin{itemize}
  \item Near-term narrative: \{\texttt{narrative\_near}\}
  \item Severity: \{\texttt{severity\_near}\}
  \item Background tendency: \{\texttt{background}\}
  \item Past-window variability: \{\texttt{variability}\}
  \item Pattern cue: \{\texttt{pattern}\}
\end{itemize}

Rules:
(1) Describe observed drivers only (e.g., inversions, wind, humidity, transport, anthropogenic sources); 
(2) No forecasts, numbers, or speculative language; 
(3) Keep factual, neutral, natural; one paragraph only.
\label{prompt_template}
\end{tcolorbox}

\begin{tcolorbox}[title={Synthetic Data Example}]
\textbf{Our Synthetic Text:}
Over the past seven days, regional air quality has shown a clear pattern of deterioration, transitioning into a period of moderate severity. This worsening was primarily driven by \textbf{limited atmospheric dispersion}, as observed through a series of weak temperature inversions that \textbf{trapped pollutants near the surface}. While background synoptic conditions remained generally mild and stable, high variability characterized the daily conditions, with significant swings in observed air quality. This high amplitude was influenced by intermittent periods of very \textbf{low wind speeds}, which allowed for pollutant accumulation, punctuated by brief episodes of \textbf{slightly improved ventilation}. A brief stagnation signal was noted, emerging without a strong, sustained background weather pattern to either reinforce or disperse it. Local observations confirmed contributions from
typical anthropogenic sources throughout this period of constrained mixing. 
\\

\textbf{Time-MMD Original (Non-NaN) Text:}

The Air Quality Health Advisory was issued for the New York City Metro region on June 25, 2022, due to high ozone levels, with an AQI value expected to exceed 100. The advisory was in effect from 11 a.m. to 11 p.m.
\label{synthetic example}
\end{tcolorbox}

\section{Additional Experiments and Discussion on MoME}
\label{app:discussion_MoME}

\subsection{Evaluation Setting for Main Experiments}
\label{app:MoME_additional_tableseter_main}
Table~\ref{tab:main_config} summarizes the key hyperparameter configurations used in the main experiments. To ensure a fair comparison, we standardize key hyperparameters, such as hidden dimension and dropout rate, across comparable models unless explicitly stated otherwise. This setting is also applied to the token-fusion baselines, including MLP (MoE)+LM, Time-MoE+LM, DLinear+LM, and the five fusion variants in Table~\ref{tab:fusion_comparison}. For methods using a pre-trained LLM or TSFM (e.g., MoME, Time-MoE+LM), the LLM/TSFM parameters are frozen throughout training, and only the task-specific prediction head is optimized. For other supervised models, all parameters are trained from scratch. All model comparisons use the same fixed data partitions. The main tables report results under the default seed, while additional multi-seed experiments with different parameter initializations are reported in Table~\ref{tab:seed} of Appendix~\ref{app:seed}. All models are implemented in PyTorch and trained on a single NVIDIA RTX A6000 GPU with 48 GB memory.


\begin{table}[]
    \centering
    \caption{Hyperparameter configuration for main experiments.}
    \resizebox{0.8\linewidth}{!}{
    \begin{tabular}{l}
\toprule
\textbf{Hyperparameter Category}  \\
\midrule
\textbf{TS-Encoder:}\\
Layers = 3; Hidden dim = 32; Dropout = 0.1; Number of Heads = 4; Patch Length = 4 \\

\textbf{Expert Modulation:}\\ 
Experts = 4; Top-$K$ = 2; Context Token = 3; EiLM = Enable/Disable; RM = Enable/Disable\\

\textbf{LLM Encoder:} \\
LLM Hidden Dimension = LLM-Dependent; Max Text Length = 2048 \\

\bottomrule
\end{tabular}
    }
    \label{tab:main_config}
\end{table}

\subsection{Robustness Evaluation}
\label{app:seed}

\begin{table}[]
    \centering
    \caption{Robustness evaluation for Table~\ref{tab:main}. Results are reported as mean $\pm$ standard deviation. Accuracy (Acc.) is reported in percentage (\%).}
    \resizebox{\linewidth}{!}{
    \begin{tabular}{lcc c cc c}
\toprule
\multirow{2}{*}{Model} &
\multicolumn{2}{c}{Finance (Long) Forecast} &
Finance (Long) Trend &
\multicolumn{2}{c}{Weather (Long) Forecast} &
Weather (Long) Trend \\
\cmidrule(lr){2-3}
\cmidrule(lr){4-4}
\cmidrule(lr){5-6}
\cmidrule(lr){7-7}
& MAPE $\downarrow$ & MAE $\downarrow$
& 3-way Acc. $\uparrow$
& MSE $\downarrow$ & MAE $\downarrow$
& Future Acc. $\uparrow$ \\
\midrule

PatchTST
& $3.919 \pm 0.088$
& $3.577 \pm 0.062$
& $49.087 \pm 1.101$
& $13.149 \pm 0.586$
& $2.754 \pm 0.054$
& $49.323 \pm 0.977$ \\

DLinear
& $3.864 \pm 0.017$
& $3.636 \pm 0.018$
& $44.292 \pm 2.913$
& $12.421 \pm 0.052$
& $2.676 \pm 0.001$
& $49.594 \pm 0.813$ \\

Time-LLM
& $4.097 \pm 0.090$
& $3.695 \pm 0.037$
& $46.461 \pm 3.430$
& $12.416 \pm 0.117$
& $2.682 \pm 0.022$
& $50.136 \pm 0.469$ \\

Time-MoE
& $4.884 \pm 0.712$
& $4.663 \pm 0.818$
& $47.603 \pm 2.470$
& $30.642 \pm 0.867$
& $4.281 \pm 0.097$
& $46.884 \pm 1.084$ \\

DLinearP+LM
& $5.667 \pm 0.072$
& $5.202 \pm 0.090$
& $47.146 \pm 1.046$
& $15.637 \pm 1.931$
& $2.999 \pm 0.172$
& $34.688 \pm 2.347$ \\

Time-MoE+LM
& $3.837 \pm 0.189$
& $3.555 \pm 0.129$
& $48.630 \pm 1.812$
& $13.405 \pm 0.221$
& $2.777 \pm 0.018$
& $44.625 \pm 5.725$ \\

\midrule
\textbf{MoME (Ours)}
& $\mathbf{3.544 \pm 0.054}$
& $\mathbf{3.305 \pm 0.020}$
& $\mathbf{63.356 \pm 0.906}$
& $\mathbf{12.321 \pm 0.610}$
& $\mathbf{2.662 \pm 0.061}$
& $\mathbf{54.923 \pm 1.634}$ \\

\bottomrule
\end{tabular}
    }
    \label{tab:seed}
\end{table}

We also conducted several experiments for our main experiments on the MT-Bench datasets. We tested three random seeds to further evaluate the robustness of MoME across forecasting and trend-prediction tasks. MoME achieves the best average performance on all evaluated metrics in both the finance and weather domains. The improvements are particularly stable for finance trend prediction, while MoME also maintains competitive variance in the forecasting tasks. More importantly, MoME consistently outperforms the fusion-based multi-modal baselines (i.e., \textit{DLinearP+LM} and \textit{Time-MoE+LM}). Overall, these results suggest that the observed gains are not sensitive to a specific random initialization and provide additional evidence for the reliability of our Expert Modulation for multi-modal learning.

\subsection{MoME on Time-MMD}
\label{app:mome,time-mmd}

We conduct additional experiments with Time-MMD (dataset introduction in Appendix~\ref{app:data3}), to further evaluate the effectiveness of MoME in different scenarios. As shown in Table~\ref{tab:timemmd}, MoME outperforms the baselines by a large margin in various domains. This suggests that when textual information is highly aligned and provides clean complementary signals, expert modulation can effectively utilize this information for time series prediction. Notably, other MMTSP models (e.g., MLP (MoE)+LM) fail to achieve similar performance, further highlighting the advantage of expert modulation in integrating text and time series.

\begin{table}[]
    \centering
    \caption{Time-MMD results and ablations. Improvement (\%) is computed relative to the $w/o$ EM variant. Health (AFR) metrics are reported in units of $10^2$.}
    \resizebox{\linewidth}{!}{
          \begin{tabular}{l *{10}{c}}
    \toprule
    \multirow{2}{*}{\textbf{Method}} &
    \multicolumn{2}{c}{\textbf{Environment}} &
    \multicolumn{2}{c}{\textbf{Energy}} &
    \multicolumn{2}{c}{\textbf{Health (US)}} &
    \multicolumn{2}{c}{\textbf{Health (AFR)}} &
    \multicolumn{2}{c}{\textbf{Social Good}} \\
    \cmidrule(lr){2-3} \cmidrule(lr){4-5} \cmidrule(lr){6-7} \cmidrule(lr){8-9} \cmidrule(lr){10-11}
     & MAPE & MAE & MSE & MAE & MSE & MAE & MSE & MAE & MSE & MAE \\
    \midrule
    PatchTST          & 28.270 & 16.331 & 0.064 & 0.187 & 1.503 & 0.832 & 0.239 & 3.465 & 2.406 & 0.786 \\
    DLinear           & 28.519 & 16.294 & 0.031 & 0.120 & 0.929 & 0.614 & 0.186 & 2.970 & 2.116 & 0.709 \\
    TS-Mix            & 27.682 & 16.162 & 0.031 & 0.123 & 0.867 & 0.553 & 0.206 & 3.106 & 2.413 & 0.654 \\
    Time-MoE          & 25.659 & 14.947 & 0.016 & 0.082 & 0.789 & 0.483 & 0.152 & 2.621 & \textbf{1.795} & \textbf{0.519} \\
    Time-LLM          & 26.712 & 15.318 & 0.016 & 0.087 & 0.851 & 0.530 & 0.141 & 2.567 & 2.020 & 0.541 \\
    GPT4TS            & 26.563 & 15.206 & 0.015 & 0.082 & 0.824 & 0.504 & \textbf{0.134} & \textbf{2.551} & 1.817 & 0.496 \\
    Time-MoE+LM       & \textbf{25.141} & \textbf{14.925} & \textcolor{red}{\textbf{0.014}} & \textcolor{red}{\textbf{0.078}} & \textbf{0.587} & \textbf{0.445} & 0.197 & 2.894 & 2.191 & 0.739 \\
    DLinearP+LM        & 25.628 & 14.942 & 0.015 & 0.079 & 0.730 & 0.468 & 0.173 & 2.682 & 1.929 & 0.744 \\
    MLP(MoE)+LM       & 26.886 & 15.520 & 0.021 & 0.098 & 0.802 & 0.489 & 0.195 & 2.992 & 1.908 & 0.730 \\
    MoME (\textbf{Ours})              & \textcolor{red}{\textbf{15.434}} & \textcolor{red}{\textbf{8.317}} & \textbf{0.015} & \textbf{0.081} & \textcolor{red}{\textbf{0.379}} & \textcolor{red}{\textbf{0.359}} & \textcolor{red}{\textbf{0.103}} & \textcolor{red}{\textbf{2.303}} & \textcolor{red}{\textbf{1.419}} & \textcolor{red}{\textbf{0.417}} \\
    \midrule
    \multicolumn{11}{l}{\textbf{Ablation:}} \\
    MoME $w/o$ EM       & 25.678 & 14.912 & 0.018 & 0.086 & 0.808 & 0.518 & 0.125 & 2.318 & 1.743 & 0.487 \\
    \midrule
    \textit{Imp. (\%)}& 39.894 & 44.226 & 16.667 & 5.814 & 53.094 & 30.695 & 17.600 & 0.647 & 18.589 & 14.374 \\
    \bottomrule
  \end{tabular}
    }
    \label{tab:timemmd}
\end{table}

\subsection{Comparison of Fusion Strategies}
\label{app:fusion-compare}

To further validate the effectiveness of our proposed expert-modulation-based modality interaction, we compare it against several widely used fusion strategies in multi-modal time series models. Recent studies~\citep{jin2024position, zhang2025does} emphasize that token-level fusion plays a critical role in integrating textual and temporal information and examine various fusion strategies that substantially affect downstream performance. Based on the principles summarized in these studies, we implement five representative baselines that cover both early-fusion and late-fusion methods. These baselines serve as comparison points for assessing whether expert-level modulation is a more robust mechanism than conventional token-fusion approaches.

The ablation models are as follows: (i) E1 [Early Fusion, Concatenation]: The time series token embeddings are first projected to the LLM's hidden dimension, concatenated with the tokenized text, and then jointly fed into the LLM backbone; (ii) E2 [Early Fusion, Cross Attention]: Temporal token embeddings are attended to raw text tokens through token-level cross attention; then the enhanced tokens are fed into the LLM backbone; (iii) L1 [Late Fusion, Additive]: The LLM processes the text independently, and its pooled embedding is projected and added to the output of the time series encoder before entering the downstream prediction head. (iv) L2 [Late Fusion, Concatenation]: It is similar to the L1 approach, except that it replaces addition with concatenation. (v) L3 [Late Fusion, Cross Attention]: Time series final embeddings are refined by cross attention over the LLM-derived text embeddings and then sent to the downstream head for prediction.

To ensure a fair comparison, all models use the same time series backbone, MoME ($w/o$\, EM), a uni-modal MLP-based MoE time series model without expert modulation. They also use the same LLM, Qwen-MoE, and Figure~\ref{fig:main} visualizes the architectures of the ablated models. The complete results for the comparative study are summarized in Table~\ref{tab:fusion_comparison}. Across the experiments, we evaluate not only downstream performance but also training speed and memory consumption. Our proposed expert-modulation method, MoME, achieves the best performance on 7 of 8 tasks while offering significantly faster training per iteration. Notably, the memory consumption of our method is also low, which is comparable to simple fusion approaches (E1, L1, L2), and is substantially smaller than all cross attention based baselines (E2 and L3). Together, these results demonstrate that the expert-modulation framework is an efficient and effective multi-modal integration mechanism compared to classical token-fusion strategies.

\begin{table}[]
    \centering
    \caption{Comparison with token-fusion strategies. E1–E2 denote early-fusion methods, and L1–L3 denote late-fusion methods. For example, MoME ($w/o$ EM) + E1 indicates an early token-fusion variant that injects textual information by token concatenation.}
    \resizebox{\linewidth}{!}{
    \definecolor{myred}{RGB}{200,0,0}
\newcommand{\red}[1]{\textcolor{myred}{\textbf{#1}}}
\newcommand{\bfr}[1]{\textbf{#1}}

\begin{tabular}{l|cccccc|cccccc}
\toprule
\multirow{3}{*}{\textbf{Model \textbackslash Metric}} & \multicolumn{6}{c|}{\textbf{MT-Finance}} & \multicolumn{6}{c}{\textbf{SocialGood}} \\
\cline{2-13}
& \multicolumn{2}{c}{\textbf{Time (s/iter)}} & \multicolumn{2}{c}{\textbf{Trainable Parameters (M)}} & \multicolumn{2}{c|}{\textbf{Performance}} & \multicolumn{2}{c}{\textbf{Time (s/iter)}} & \multicolumn{2}{c}{\textbf{Trainable Parameters (M)}} & \multicolumn{2}{c}{\textbf{Performance}} \\
\cline{2-13}
& Train & Inference & TS-Encoder & Instr/Align & 3-Acc & 5-Acc & Train & Inference & TS-Encoder & Instr/Align & MSE & MAE \\
\midrule
\textbf{Uni-Modal Baseline}\\
MoME($w/o$ EM)& 0.01 & 0.01- & 0.0095 & \textbackslash & 46.233 & 40.411 & 0.01- & 0.01- & 0.0095 & \textbackslash & 1.743 & 0.487 \\
\textbf{Expert Modulation}\\
MoME    & 0.47 & 0.46 & 0.0126 & 0.0700 & \red{63.699} & \red{51.027} & 0.38 & 0.36 & 0.0126 & 0.0700 & \red{1.419} & \red{0.417} \\
\textbf{Token Alignment}\\
MoME($w/o$ EM)+E1 & 1.19 & 0.47 & 0.0095 & 0.0676 & \bfr{48.288} & 33.219 & 1.03 & 0.37 & 0.0095 & 0.0676 & 1.908 & 0.730 \\
MoME($w/o$ EM)+E2 & 1.25 & 0.47 & 0.0095 & 16.853 & \bfr{47.603} & \bfr{41.096} & 1.05 & 0.37 & 0.0095 & 16.853 & \bfr{1.504} & 0.534 \\
MoME($w/o$ EM)+L1 & 1.19 & 0.47 & 0.0095 & 0.0676 & \bfr{63.013} & \red{51.027} & 1.03 & 0.36 & 0.0095 & 0.0676 & 2.125 & 0.620 \\
MoME($w/o$ EM)+L2 & 1.21 & 0.47 & 0.0095 & 0.0676 & \bfr{62.329} & \bfr{47.603} & 1.03 & 0.37 & 0.0095 & 0.0676 & \bfr{1.699} & 0.553 \\
MoME($w/o$ EM)+L3 & 1.58 & 0.47 & 0.0095 & 16.853 & \bfr{49.315} & \bfr{41.781} & 0.49 & 0.37 & 0.0095 & 16.853 & 1.869 & 0.525 \\

\midrule
\midrule

\multirow{3}{*}{\textbf{Model \textbackslash Metric}} & \multicolumn{6}{c|}{\textbf{MT-Weather}} & \multicolumn{6}{c}{\textbf{Health-US}} \\
\cline{2-13}
& \multicolumn{2}{c}{\textbf{Time (s/iter)}} & \multicolumn{2}{c}{\textbf{Trainable Parameters (M)}} & \multicolumn{2}{c|}{\textbf{Performance}} & \multicolumn{2}{c}{\textbf{Time (s/iter)}} & \multicolumn{2}{c}{\textbf{Trainable Parameters (M)}} & \multicolumn{2}{c}{\textbf{Performance}} \\
\cline{2-13}
& Train & Inference & TS-Encoder & Instr/Align & P-Acc & F-Acc & Train & Inference & TS-Encoder & Instr/Align & MSE & MAE \\
\midrule
\textbf{Uni-Modal Baseline}\\
MoME($w/o$ EM) & 0.01- & 0.01- & 0.0095 & \textbackslash & 87.534 & 49.594 & 0.01- & 0.01- & 0.0095 & \textbackslash & 0.808 & 0.518 \\
\textbf{Expert Modulation}\\
MoME     & 0.49 & 0.47 & 0.0126 & 0.0700 & \red{93.496} & \red{53.388} & 0.37 & 0.36 & 0.0126 & 0.0700 & \red{0.379} & \red{0.359} \\

\textbf{Token Alignment}\\
MoME($w/o$ EM)+E1 & 1.22 & 0.48 & 0.0095 & 0.0676 & 86.450 & 41.192 & 1.04 & 0.37 & 0.0095 & 0.0676 & \bfr{0.802} & \bfr{0.489} \\
MoME($w/o$ EM)+E2 & 1.27 & 0.48 & 0.0095 & 16.853 & \bfr{87.805} & \bfr{49.594} & 1.05 & 0.38 & 0.0095 & 16.853 & \bfr{0.772} & 0.562 \\
MoME($w/o$ EM)+L1 & 1.21 & 0.47 & 0.0095 & 0.0676 & 64.228 & 38.482 & 1.05 & 0.37 & 0.0095 & 0.0676 & 1.694 & 0.872 \\
MoME($w/o$ EM)+L2 & 1.22 & 0.48 & 0.0095 & 0.0676 & 63.415 & 44.986 & 1.05 & 0.38 & 0.0095 & 0.0676 & 1.022 & 0.693 \\
MoME($w/o$ EM)+L3 & 1.61 & 0.47 & 0.0095 & 16.853 & \bfr{90.515} & 42.818 & 0.51 & 0.38 & 0.0095 & 16.853 & 1.088 & 0.663 \\

\bottomrule
\end{tabular}

    }
    \label{tab:fusion_comparison}
\end{table}

\subsection{Training Dynamics}
\begin{figure}
    \centering
    \includegraphics[width=0.49\linewidth]{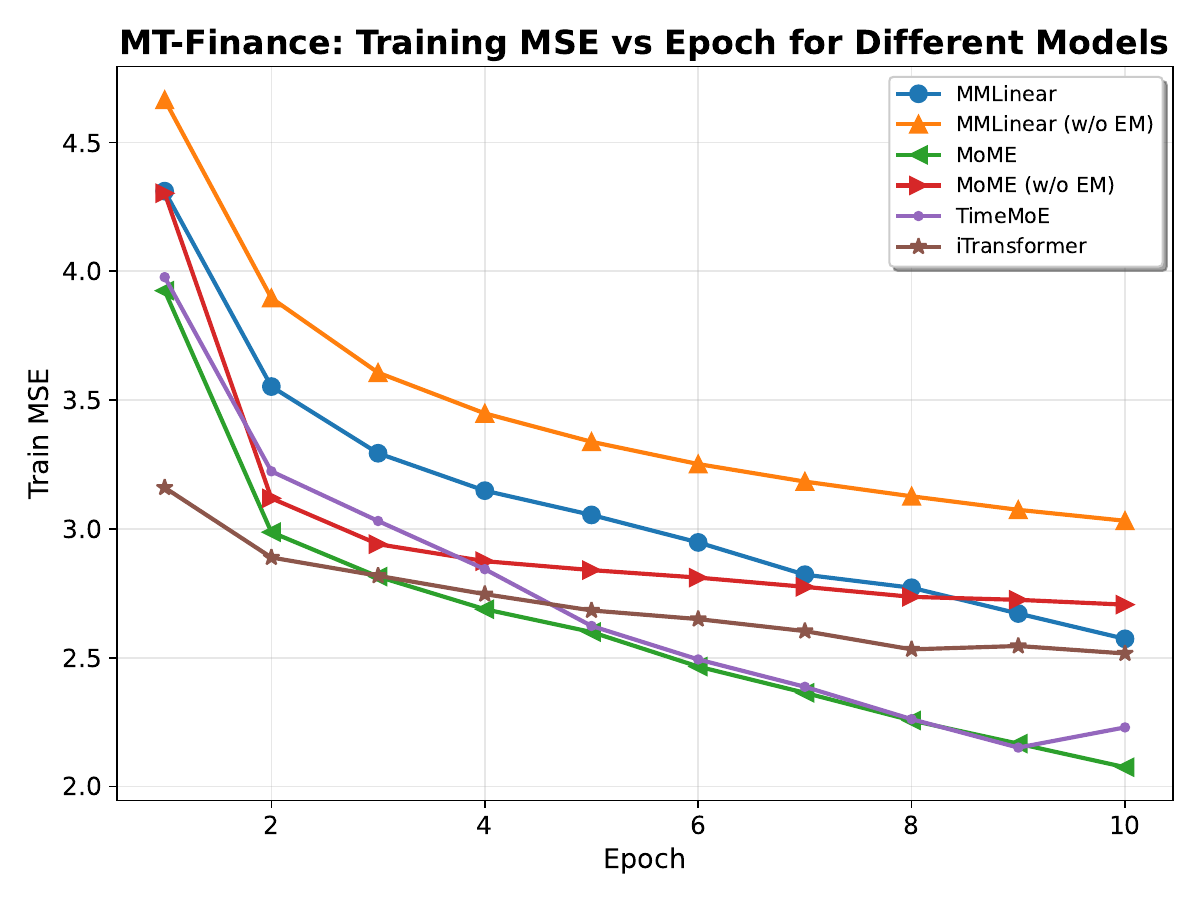}
    \hfill 
    \includegraphics[width=0.49\linewidth]{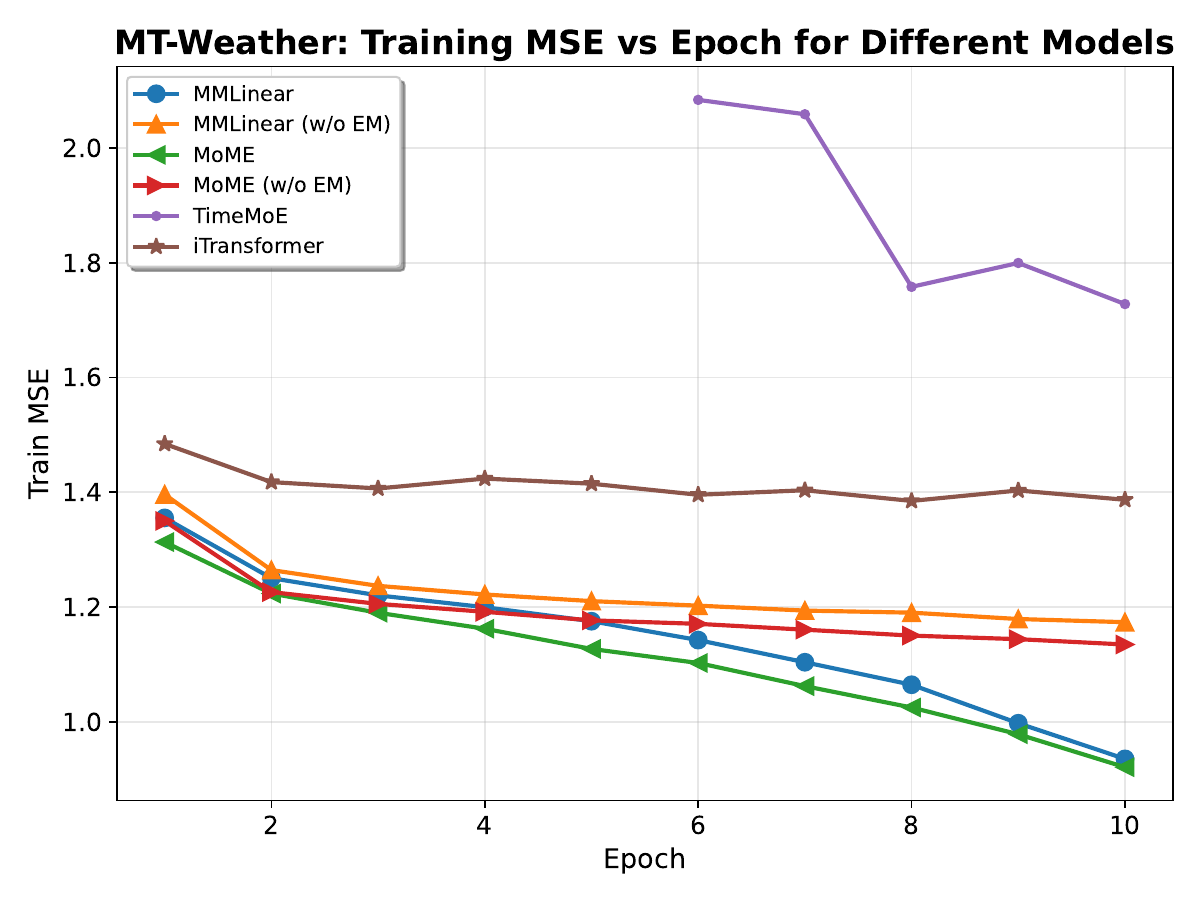}
    \caption{Comparison for training efficiency on two benchmarks.}
    \label{fig:loss}
\end{figure}

Figure~\ref{fig:loss} compares the training dynamics of different models on the MT-Bench dataset. MoME and MMLinear exhibit faster empirical convergence than iTransformer and TimeMoE. When expert modulation is removed (i.e., when the \textit{EiLM} and \textit{RM} modules are removed), the corresponding uni-modal variants, MoME ($w/o$ EM) and MMLinear ($w/o$ EM), generally converge more slowly and reach higher training losses. These observations suggest that expert modulation may facilitate model optimization in the evaluated settings, in addition to improving predictive performance. We further observe that MoME continues to reduce its training loss after several baseline models have largely plateaued, indicating that it may continue to benefit from the multi-modal training signal over a longer optimization trajectory.

\subsection{Additional Results for Main Content}
\label{app:MoME_additional_tables}

This section presents the complete numerical results underlying the figures in the main text. Table~\ref{tab:mome_hyperparameter} corresponds to Figure~\ref{fig:hyperparameter_fig}, and Table~\ref{tab:vary_llm} corresponds to Figure~\ref{fig:vary_llm}.

\begin{table}[]
    \centering
    \caption{Ablation on how the number of experts, activated experts, and LLM-derived context tokens jointly influence proposed model performance. Here, the ``CT-$k$'' stands for $k$ context tokens extracted by LLM, and the rows vary along two dimensions: the total number of experts in the MoE module (\#Experts), and the number of experts activated by the Top-$K$ routing.}
    \resizebox{\linewidth}{!}{
    \begin{tabular}{cccccccccccccccc}
\toprule
\multirow{2}{*}{\# Experts} & \multirow{2}{*}{Activated} &
\multicolumn{2}{c}{CT-1} &
\multicolumn{2}{c}{CT-2} &
\multicolumn{2}{c}{CT-3} &
\multicolumn{2}{c}{CT-4} &
\multicolumn{2}{c}{CT-5} &
\multicolumn{2}{c}{CT-6} &
\multirow{2}{*}{Avg\_MAPE} & 
\multirow{2}{*}{Avg\_MAE} \\
\cmidrule(lr){3-14}
& & MAPE & MAE & MAPE & MAE & MAPE & MAE & MAPE & MAE & MAPE & MAE & MAPE & MAE & & \\
\midrule
3 & 1 & 3.672 & 3.432 & 3.572 & 3.357 & \textbf{3.685} & 3.426 & 3.704 & 3.363 & 3.765 & 3.392 & 3.679 & 3.394 & 3.680 & 3.394 \\
3 & 2 & \textbf{3.621} & \textbf{3.299} & \textbf{3.530} & \textbf{3.243} & 3.690 & 3.445 & 3.687 & 3.501 & 3.574 & 3.299 & 3.621 & 3.233 & 3.621 & 3.337 \\
3 & 3 & 3.627 & 3.345 & 3.587 & 3.400 & 3.726 & \textbf{3.379} & \textbf{3.489} & \textbf{3.309} & \textbf{3.515} & \textbf{3.259} & \textbf{3.513} & \textbf{3.194} & 3.576 & 3.343 \\
\midrule
4 & 1 & 3.746 & 3.488 & 3.750 & 3.513 & 3.631 & 3.382 & \textbf{3.632} & 3.342 & 3.683 & 3.395 & 3.650 & 3.360 & 3.678 & 3.397 \\
4 & 2 & \textbf{3.461} & \textbf{3.139} & 3.705 & 3.469 & 3.706 & 3.415 & 3.682 & \textbf{3.338} & 3.649 & 3.329 & 3.646 & \textbf{3.290} & 3.641 & 3.330 \\
4 & 3 & 3.687 & 3.583 & 3.632 & 3.390 & \textbf{3.561} & \textbf{3.242} & 3.676 & \textbf{3.338} & \textbf{3.572} & \textbf{3.331} & \textbf{3.601} & 3.339 & 3.621 & 3.370 \\
4 & 4 & 3.647 & 3.397 & \textbf{3.563} & \textbf{3.229} & 3.630 & 3.305 & 3.740 & 3.428 & 3.596 & 3.303 & 3.783 & 3.422 & 3.659 & 3.347 \\
\midrule
5 & 1 & \textbf{3.572} & \textbf{3.325} & 3.856 & 3.489 & 3.634 & 3.384 & 3.691 & 3.272 & 3.704 & 3.383 & 3.871 & 3.487 & 3.721 & 3.390 \\
5 & 2 & 3.710 & 3.445 & 3.758 & 3.590 & 3.800 & 3.481 & 3.617 & 3.324 & \textbf{3.584} & \textbf{3.290} & \textbf{3.556} & 3.287 & 3.671 & 3.341 \\
5 & 3 & 3.680 & 3.335 & 3.658 & 3.326 & 3.596 & 3.302 & 3.635 & 3.290 & 3.651 & 3.332 & 3.652 & 3.347 & 3.645 & 3.320 \\
5 & 4 & 3.596 & 3.436 & 3.594 & 3.370 & \textbf{3.477} & \textbf{3.174} & \textbf{3.594} & \textbf{3.266} & 3.710 & 3.401 & 3.587 & \textbf{3.231} & 3.593 & 3.296 \\
5 & 5 & 3.597 & 3.426 & \textbf{3.531} & \textbf{3.300} & 3.551 & 3.336 & 3.597 & 3.356 & 3.708 & 3.425 & 3.696 & 3.429 & 3.613 & 3.379 \\
\midrule
6 & 2 & 3.721 & 3.340 & \textbf{3.627} & \textbf{3.316} & 3.653 & 3.340 & 3.669 & 3.351 & 3.648 & 3.324 & \textbf{3.596} & \textbf{3.288} & 3.652 & 3.326 \\
6 & 4 & \textbf{3.559} & \textbf{3.293} & 3.686 & 3.476 & \textbf{3.611} & \textbf{3.249} & 3.651 & 3.320 & 3.742 & 3.408 & 3.653 & 3.305 & 3.650 & 3.341 \\
6 & 6 & 3.572 & 3.357 & 3.659 & 3.520 & 3.884 & 3.676 & \textbf{3.504} & \textbf{3.182} & \textbf{3.596} & \textbf{3.291} & 3.883 & 3.574 & 3.683 & 3.433 \\
\midrule
\textbf{Avg} & & 3.631 & 3.376 & 3.647 & 3.393 & 3.656 & 3.369 & 3.638 & 3.332 & 3.646 & 3.343 & 3.666 & 3.345 &  &  \\
\bottomrule
\end{tabular}
    }
    \label{tab:mome_hyperparameter}
\end{table}

\subsection{Dynamics of Token Routing}
\label{app:expert-select}

While Section~\ref{sec:Ablation} evaluates the downstream effect of different modulation mechanisms, here we further examine how expert modulation influences routing dynamics. Prior MoE studies suggest that similar tokens are often routed to similar subsets of experts, leading to expert specialization. We also observe this behavior in uni-modal MoE-based time series models (Appendix~\ref{app:uni_routing}). In this section, we further investigate whether such specialization is preserved when external textual information is incorporated through expert modulation. Specifically, we visualize routing dynamics on the Finance and SocialGood datasets under MoME model.

Based on our observations, routing similarity in time series is mainly associated with two aspects: (i) \textit{morphological patterns}, which describe the shape of local temporal dynamics, and (ii) \textit{magnitudes}, which describe the overall signal scale.

\begin{figure}
    \centering
    \includegraphics[width=\linewidth]{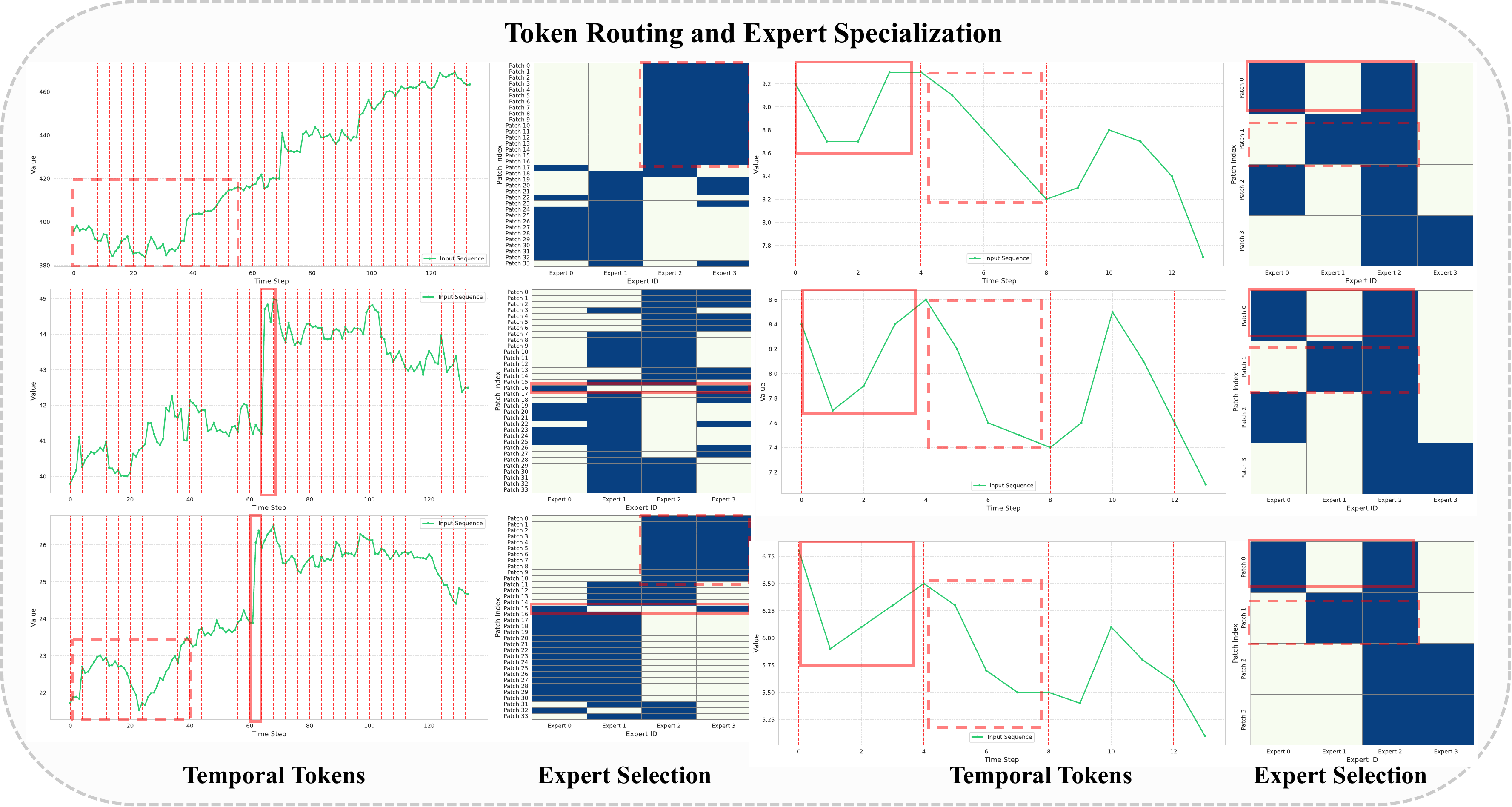}
    \caption{The dynamic of routing when router modulation activated.}
    \label{fig:expert_selection_magnitude}
\end{figure}

\paragraph{Magnitude-Based Selection.}
One form of specialization is shown in the left panel of Figure~\ref{fig:expert_selection_magnitude}. We observe that low-magnitude patches are usually routed to Experts 2 and 3, while high-magnitude patches tend to activate Experts 1 and 2. In addition, we observe that two special patches with abrupt upward movements are assigned to Experts 1 and 4. This suggests that the router can also capture sharp local changes beyond simple magnitude grouping.

\begin{table}[]
    \centering
    \caption{Evaluation of MoME using different LLM backbones.}
    \resizebox{0.9\linewidth}{!}{
    \begin{tabular}{lcccccc}
\toprule
\multirow{2}{*}{Model}
& \multicolumn{2}{c}{Finance Forecast}
& \multicolumn{2}{c}{Weather Forecast}
& \multicolumn{1}{c}{Finance Trend}
& \multicolumn{1}{c}{Weather Trend} \\
\cmidrule(lr){2-3}
\cmidrule(lr){4-5}
\cmidrule(lr){6-6}
\cmidrule(lr){7-7}
& MAPE$\downarrow$
& MAE$\downarrow$
& MSE$\downarrow$
& MAE$\downarrow$
& ACC$\uparrow$
& ACC$\uparrow$ \\
\midrule
PatchTST
& 3.832 & 3.604 & 14.312 & 2.857 & 50.802 & 26.558 \\

MoME (DeepSeek-16B)
& 3.590 & 3.241 & \textbf{11.898} & \textbf{2.623} & 60.959 & 50.948 \\

MoME (MiniMax-32B)
& 3.559 & 3.207 & 12.054 & 2.642 & 64.041 & 48.239 \\

MoME (Qwen3-14B)
& 3.628 & 3.337 & 12.353 & 2.663 & \textbf{67.123} & 49.323 \\

MoME (Qwen3.5-4B)
& \textbf{3.508} & \textbf{3.179} & 12.611 & 2.717 & 65.411 & 49.051 \\

MoME (Qwen3.6-27B)
& 3.687 & 3.355 & 12.758 & 2.719 & 66.096 & 53.117 \\

MoME (Qwen3.6-35B-A3B)
& 3.765 & 3.392 & 12.248 & 2.640 & 61.301 & \textbf{54.743} \\
\bottomrule
\end{tabular}
    }
    \label{tab:vary_llm}
\end{table}

\paragraph{Morphology-Based Selection.}
Another form of specialization is shown in the right panel of Figure~\ref{fig:expert_selection_magnitude}. We observe that $V$-shaped patches are more likely to be routed to Experts 0 and 2, while decreasing $\backslash$-shaped patches tend to activate Experts 1 and 2. This indicates that the router can distinguish local temporal morphologies and assign them to specialized expert subsets. Such morphology-based specialization is consistent with our observations in the uni-modal setting (Appendix~\ref{app:uni_routing}).

Overall, these results show that \textbf{expert modulation does not destroy the intrinsic specialization behavior of MoE}. Instead, MoME preserves meaningful routing structures based on both temporal morphology and signal magnitude. This supports the view that expert modulation complements the native advantages of MoE, allowing external textual information to influence prediction while maintaining specialized expert assignments.


\section{Case Study: How Text Helps?}
\label{app:case study}
Although Table~\ref{tab:main} demonstrates the performance gains from multi-modality data and the expert-modulation framework, these numbers alone do not reveal how textual information actually contributes to prediction. Second, it remains unclear whether the improvements stem merely from the expressive power of the LLM, or whether the model truly ``understands or comprehends'' the news content. Finally, real-world text is often imperfect or even inconsistent with future outcomes; for example, financial reports can contain conflicting assessments or incorrect projections about the prospects of a company. Understanding how MoME behaves under such ambiguous or unreliable textual input is therefore essential. To provide a deeper understanding, we conduct case studies on strong baselines and ablated variants.

\begin{figure}[]
    \centering
    \includegraphics[width=0.9\textwidth]{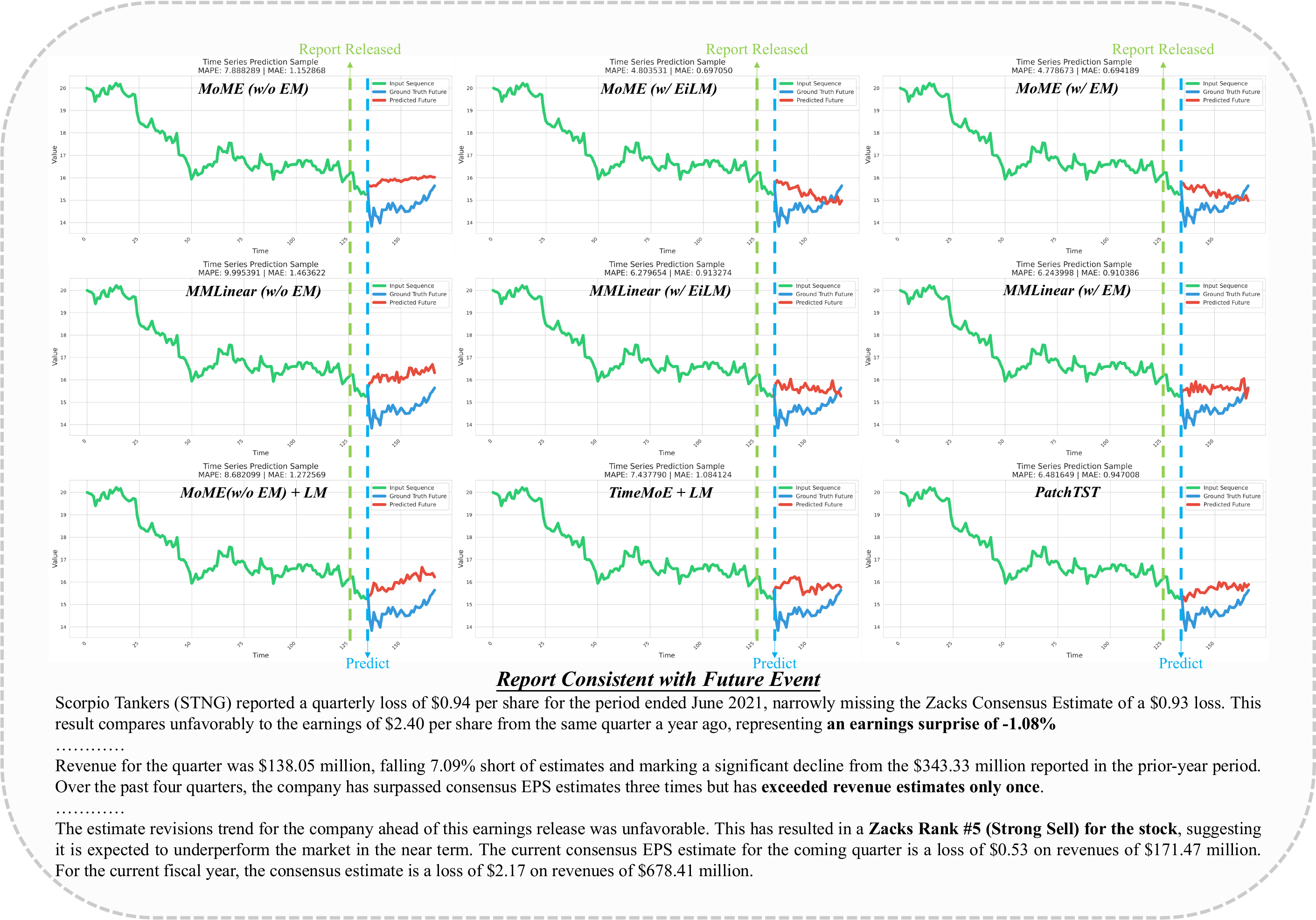}
    \caption{Stock price forecasting(1): when news report are consistent with actual future state.}
    \label{fig:FF-Case28}
\end{figure}

\begin{figure}[]
    \centering
    \includegraphics[width=0.9\textwidth]{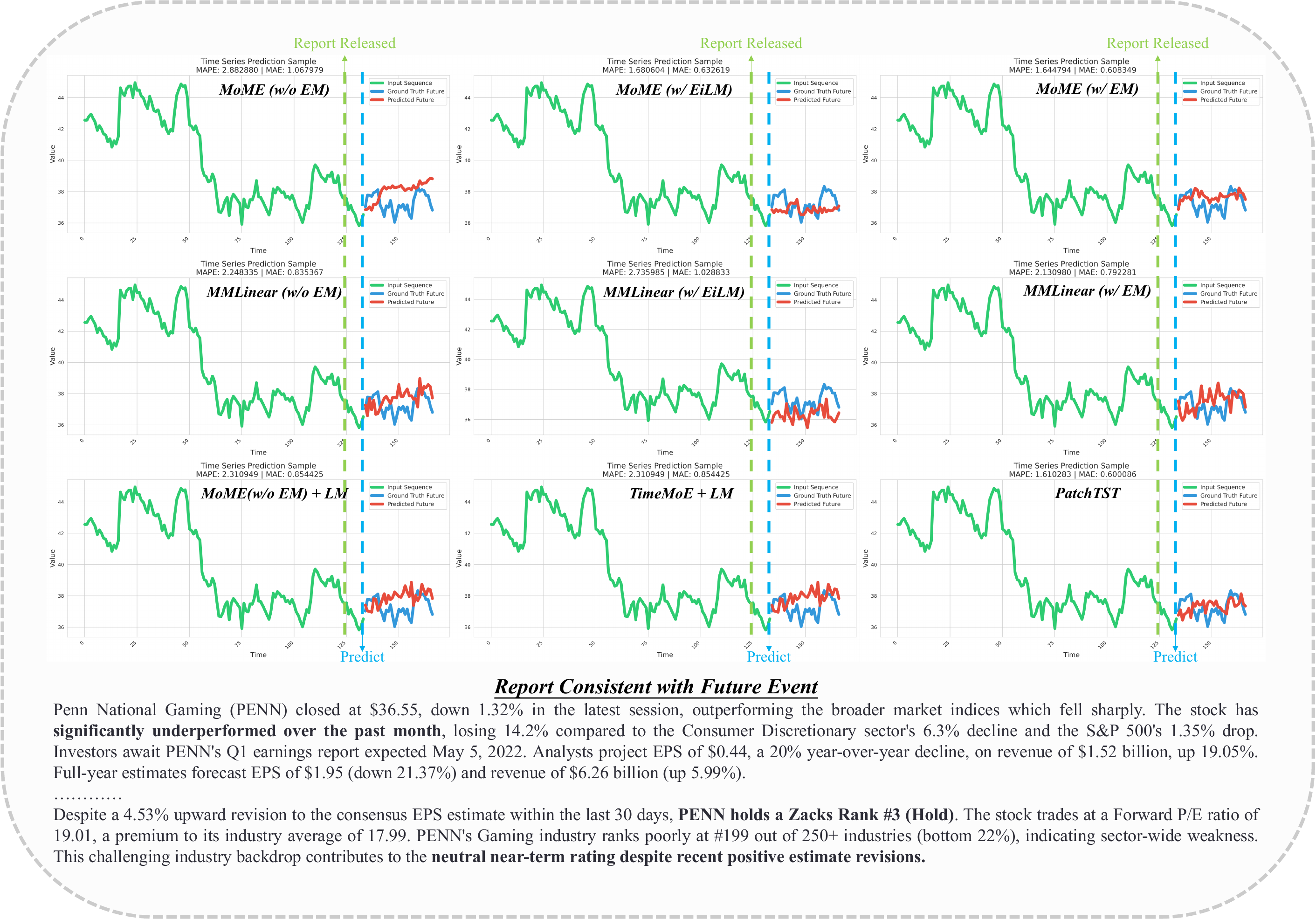}
    \caption{Stock price forecasting(2): when news report are consistent with actual future state.}
    \label{fig:FF-Case77}
\end{figure}

\paragraph{Successful Cases.}
Figures~\ref{fig:FF-Case28} and~\ref{fig:FF-Case77} present two examples in which financial reports are correct about subsequent market behavior. In Figure~\ref{fig:FF-Case28}, the report sends several negative signals, including a ``surprising negative earnings warning'' and a ``clear recommendation to sell the stock''. In this scenario, both proposed model families, [MoME ($w/o$\,EM), MoME ($w/$\,EiLM), MoME ($w/$\,EM)] and [MMLinear($w/o$\,EM),  MMLinear, MMLinear($w/$\,EM)], show clear improvements when expert modulation or router modulation is activated. Crucially, the uni-modal MoE variants [MoME($w/o$\,EM) and MMLinear($w/o$\,EM)] incorrectly predict an upward future trend. Once expert modulation is enabled, however, the predicted trajectory reverses: both models correctly forecast a downward trend, which is consistent with the negative sentiment in the report. This demonstrates that the activation of expert modulation does not merely refine the output but can fundamentally alter the predicted trend evolution. Meanwhile, early-fusion models, such as MLP (MoE)+LM and TimeMoE+LM, fail to produce the correct downward prediction, as they still forecast a positive or weakly increasing trend. This contrast highlights that using simple early fusion strategies to connect time series models and LLM has fundamental limitations.

Figure~\ref{fig:FF-Case77} provides another successful example. This report acknowledges a recent upward revision in the stock price, but ultimately concludes that the stock is expected to remain neutral in the near-term future. This subtle combination of short-term momentum and medium-term stabilization poses a significant challenge for uni-modal time series models. In contrast, our expert-modulation models, MoME ($w/$\,EiLM) and MoME ($w/$\,EM), predict a nearly flat and stable trajectory, closely matching the interpretation of the report. Together, these case studies demonstrate that our proposed modality interaction framework enables the model to effectively integrate information across modalities, producing forecasts that reflect the semantic signals embedded in both the textual reports and the time series patterns.

\begin{figure}[]
    \centering
    \includegraphics[width=0.9\linewidth]{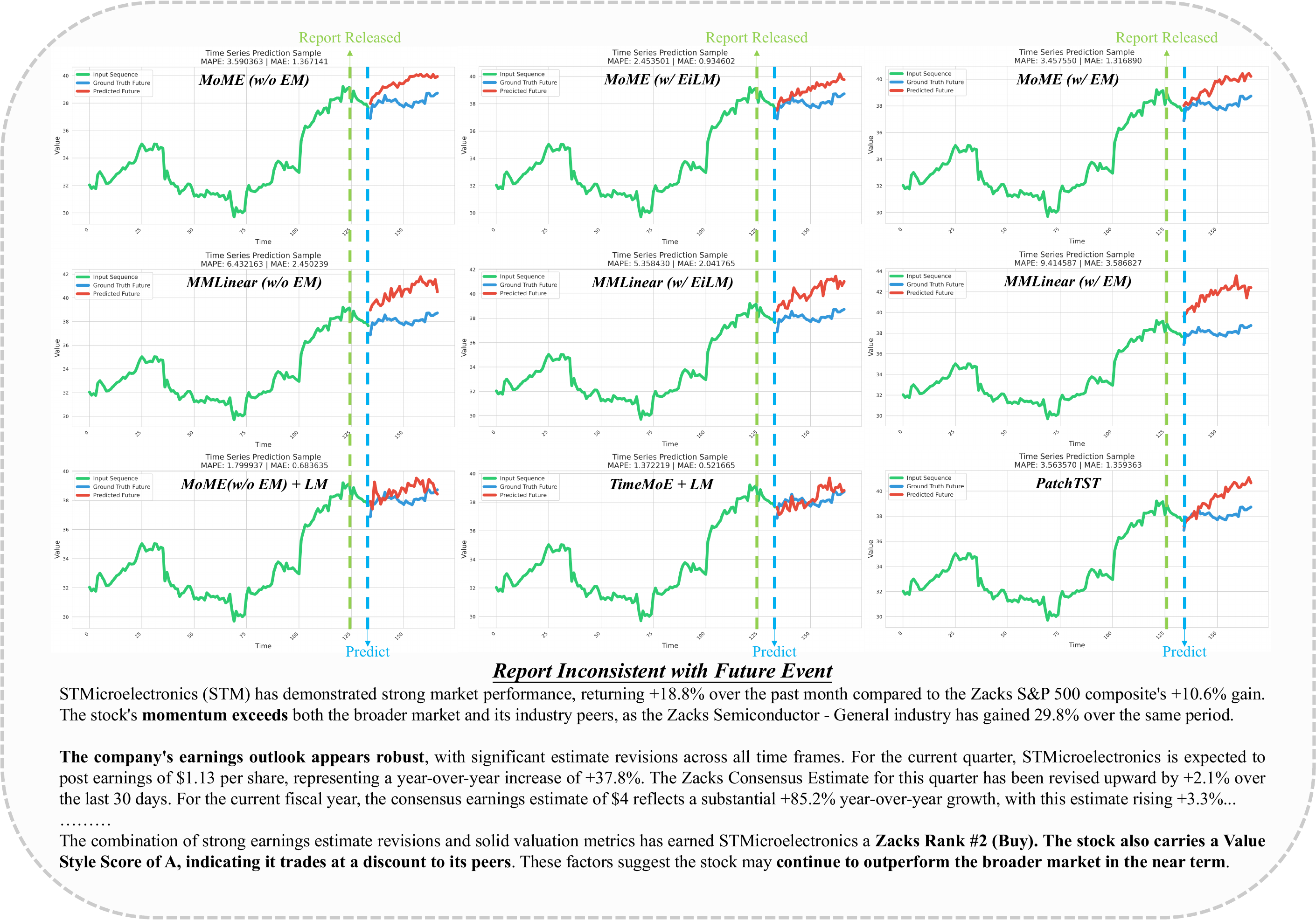}
    \caption{Stock price forecasting(3): when news report are inconsistent with actual future state.}
    \label{fig:FF-Case8}
\end{figure}

\paragraph{Struggling Cases.}
Figure~\ref{fig:FF-Case8} illustrates a clear failure case. The corresponding report expresses an overwhelmingly positive outlook for the stock, using phrases such as “the company’s earnings outlook appears to be robust” and indicating expectations that the firm will “continue to outperform the broader market in the near term.”  In this scenario, both uni-modal and multi-modal models consistently predict a rising trend, particularly, no model successfully resists the misleading reports information. This case exposes a limitation of the current framework: strongly misleading textual signals may bias the multi-modal predictor toward an incorrect trend.

\begin{figure}[]
    \centering
    \includegraphics[width=0.9\linewidth]{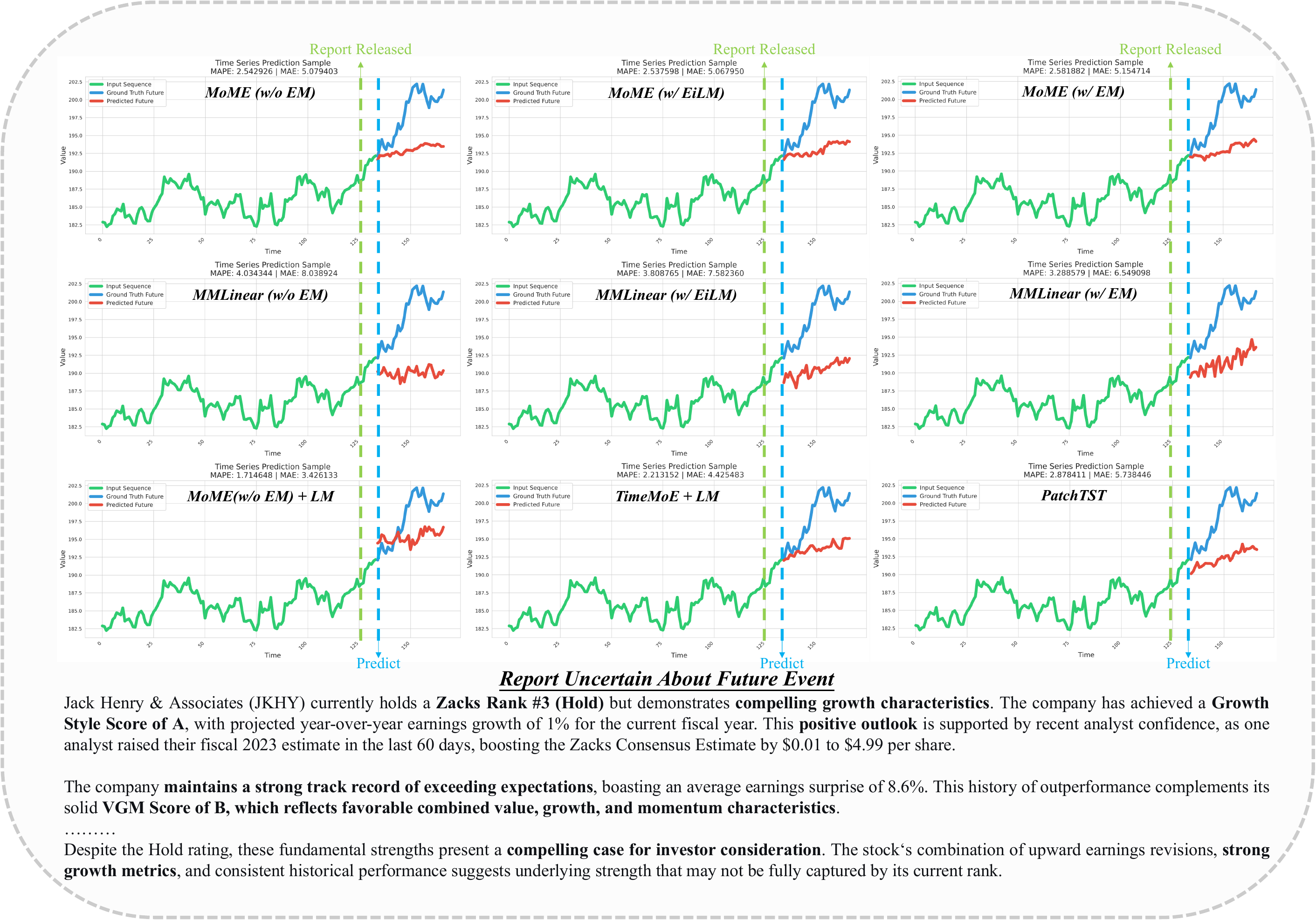}
    \caption{Stock price forecasting(4): when news report are uncertain about actual future state.}
    \label{fig:FF-Case10}
\end{figure}

Figure~\ref{fig:FF-Case10} presents another scenario in which the associated financial report conveys mixed signals: it cautiously recommends holding the stock while simultaneously highlighting a compelling growth outlook. Such contradictory cues also lead to uncertainty in model predictions. Within the group [MoME ($w/o$\,EM), MoME ($w/$\,EiLM), MoME ($w/$\,EM)], we observe that the option of expert modulation has a minimal effect, where both the trend and quantitative metrics remain largely unchanged. However, in the [MMLinear ($w/o$\,EM), MMLinear ($w/$\,EiLM), MMLinear ($w/$ EM)] group, expert modulation drastically alters the behavior of the model: without modulation, the model produces an almost flat forecast, while with expert modulation, it clearly predicts an upward trend. This contrast suggests that, under the multi-modal learning framework, the model is indeed leveraging textual information rather than relying solely on the expressive capacity of LLM. Nevertheless, when the text itself is ambiguous or contradictory, the model may still fail to form reliable forecasts.

\newpage


\newpage


\end{document}